\documentclass{article}

\PassOptionsToPackage{numbers, compress}{natbib}
\usepackage[preprint]{neurips_2024}
\usepackage[numbers]{natbib}

\usepackage[utf8]{inputenc} 
\usepackage[T1]{fontenc}    
\usepackage{hyperref}       
\usepackage{url}            
\usepackage{booktabs}       
\usepackage{amsfonts}       
\usepackage{nicefrac}       
\usepackage{microtype}      
\usepackage{multirow}
\usepackage{graphicx}
\usepackage[table]{xcolor} 
\definecolor{lightgray}{rgb}{0.7, 0.7, 0.7} 

\usepackage{caption}

\usepackage{amsmath}  
\usepackage{array}    
\usepackage{wrapfig} 
\usepackage{hhline}
\usepackage{colortbl} 
\usepackage{lineno}
\definecolor{mycolor}{RGB}{120,179,219}
\usepackage{authblk}

\title{Evaluation of Text-to-Video Generation Models: A Dynamics Perspective}

\author{\textbf{Mingxiang Liao}\textsuperscript{1}\thanks{~Equal contribution. $\dagger$ Corresponding Author.}  \quad 
\textbf{Hannan Lu}\textsuperscript{2}$^*$ \quad
\textbf{Xinyu Zhang}\textsuperscript{3,4}$^*$ \quad
\textbf{Fang Wan}\textsuperscript{1} \quad
\textbf{Tianyu Wang}\textsuperscript{1} \newline
\textbf{Yuzhong Zhao}\textsuperscript{1} \quad
\textbf{Wangmeng Zuo}\textsuperscript{2} \quad
\textbf{Qixiang Ye}\textsuperscript{1}$^\dagger$ \quad
\textbf{Jingdong Wang}\textsuperscript{4} \quad

\textsuperscript{1}University of Chinese Academy of Sciences \quad \textsuperscript{2}Harbin Institute of Technology \newline
\textsuperscript{3}The University of Adelaide \quad 
\textsuperscript{4}Baidu Inc.
}

\begin{document}

\maketitle

\begin{abstract}

Comprehensive and constructive evaluation protocols play an important role in the development of sophisticated text-to-video (T2V) generation models. 
Existing evaluation protocols primarily focus on temporal consistency and content continuity, yet largely ignoring the dynamics of video content. 
Dynamics are an essential dimension for measuring the visual vividness and the honesty of video content to text prompts. 
In this study, we propose an effective evaluation protocol, termed DEVIL, which centers on the \textit{dynamics dimension} to evaluate T2V models. 
For this purpose, we establish a new benchmark comprising text prompts that fully reflect multiple dynamics grades, and define a set of dynamics scores corresponding to various temporal granularities to comprehensively evaluate the dynamics of each generated video.
Based on the new benchmark and the dynamics scores, we assess T2V models with the design of three metrics: \textit{dynamics range}, \textit{dynamics controllability}, and \textit{dynamics-based quality}. 
Experiments show that DEVIL achieves a Pearson correlation exceeding 90\% with human ratings, demonstrating its potential to advance T2V generation models. Code is available at \href{https://github.com/MingXiangL/DEVIL}{\color{magenta}github.com/MingXiangL/DEVIL}.
\end{abstract}

\section{Introduction}

With the rapid progress of video generation technology, the demand of comprehensively evaluating model performance continues to grow.
Recent benchmarks~\cite{liu2023evalcrafter,huang2023vbench} have included various metrics, $e.g.$, generation quality, video-text alignment degree, and video content continuity, to evaluate text-to-video (T2V) generation models.
Despite the great efforts made, a fundamental characteristic of video—\textit{dynamics} remains overlooked.

Dynamics refers to the degree of visual change and interaction in the content of videos over time, encompassing object motion, action diversity, scene transitions, $etc.$ 
It is a crucial index for evaluating video generation models for the following two reasons:
$(i)$ 
Dynamics of generated video content should be honest to text prompts in practical applications. For example, it is expected that dramatic text prompts result in videos with high dynamics. 
$(ii)$ 
%
Generated videos usually show negative correlations between dynamics and quality scores~\cite{huang2023vbench, liu2023evalcrafter}, $i.e.,$ videos with higher dynamics tend to receive lower quality scores.
This allows T2V models to ``cheat'' to achieve high-quality scores by generating low-dynamic video content in many cases.

\begin{figure}
  \begin{center}
    \includegraphics[width=1.0\linewidth]{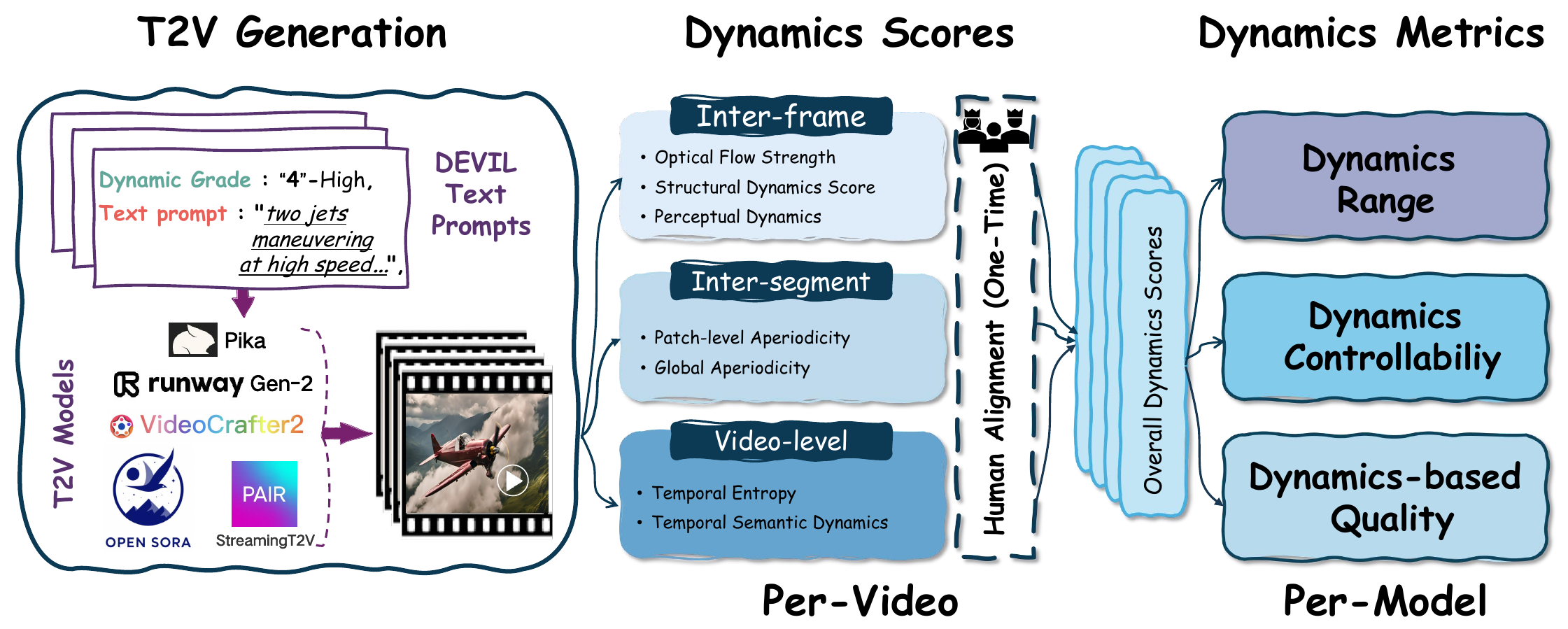}
  \end{center}
  \caption{Flowchart to calculate dynamics metrics based on dynamics scores and text prompts.}
  \label{fig:flowchat}
\end{figure}

To fully reveal the dynamics of generated videos, in this paper, we introduce a new evaluation protocol, named DEVIL.
DEVIL treats \textit{dynamics} as the primary dimension for evaluating the performance of 
T2V models.
Here, we consider three types of metrics to represent dynamics:
(i) \textit{Dynamics Range}, which measures the extent of variations in video content that the model can generate;
(ii) \textit{Dynamics Controllability}, which assesses the model's ability to manipulate video dynamics in response to text prompts;
and (iii) \textit{Dynamics-based Quality}, which evaluates the visual quality of videos with varying dynamics generated by the model.

To produce the evaluation, we first establish a benchmark comprising text prompts categorized by multiple dynamics grades.
These text prompts are collected from commonly used datasets~\cite{bain2021frozen, anne2017localizing, xu2016msr, wang2024vidprom}  and categorized according to their dynamics using a Large Language Model (LLM), GPT-4~\cite{openai2023chatgpt}, followed by further manual refinement.  
Based on the constructed text prompt benchmark, we calculate an overall dynamic score for each generated video, which is defined as a weighted sum of a series of dynamics scores at different temporal granularities.

The prompt benchmark and the overall dynamics scores of all generated videos are then utilized to evaluate T2V models with three dynamics metrics. 
This evaluation goes beyond simply maximizing dynamics scores for each video; 
it emphasizes the model's ability to produce high-quality videos across various dynamics following the instructions from text prompts.
($i$) \textit{Dynamics Range} is calculated as the range of dynamics scores for all generated videos, indicating the ability of T2V models to generate videos with both subtle and dramatic temporal variations.
($ii$) For \textit{Dynamics Controllability}, we adopt a ranking consistency-based methodology to check whether the dynamics scores of generated videos align with the dynamics of their corresponding text prompts.
($iii$) \textit{Dynamics-based Quality} is defined by integrating several quality metrics with dynamics scores. It avoids biases caused by negative correlations between video dynamics and video quality~\cite{huang2023vbench,liu2023evalcrafter}, resulting in a more comprehensive evaluation of video quality.
Finally, noting that video naturalness decreases with increasing dynamics, we also propose a naturalness metric based on a multimodal large language model, $i.e.$ Gemini-1.5 Pro~\cite{gemini2024}.

%
Upon DEVIL, we evaluate and revisit the state-of-the-art T2V models, and find:
(i) Existing datasets have biased dynamics distribution, resulting in that current generation models (especially top-ranking models like GEN-2~\cite{gen2}) typically generate slow-motion videos to obtain high quality scores.
(ii) Existing training datasets have biased text prompts on dynamics.
Training on this prompts will inevitably limit the dynamics controllability of T2V models.
(iii) Through the statistical analyses of dynamics scores, especially the naturalness metric score, existing methods display  limited real-world simulation ability.
Based on these finds, we believe, a more elaborate training data with better methods will improve the T2V performance on both quality and dynamics scores.

In summary, our contributions are:

\begin{enumerate}

\item 
We propose a novel evaluation protocol, termed DEVIL, which benchmarks T2V generation models by integrating dynamics metrics. Together with existing evaluation metrics, DEVIL builds a more comprehensive evaluation protocol.

\item
We establish a new text prompt benchmark $w.r.t.$ dynamics grades as well as a set of metrics to evaluate video dynamics across temporal granularities, facilitating the assessment of dynamics range, dynamics controllability, and dynamics-based quality.

\item 
Extensive evaluation of existing T2V generation models allows us to thoroughly analyze the capabilities of T2V models through the proposed protocol and benchmarks. The results would inspire sophisticated T2V generation methods.

\end{enumerate}

\section{Related Work}
\subsection{Text-to-Video Generation Model}
As a recent breakthrough in artificial intelligence, diffusion models have pushed video generation technology to a new height.
%
Earlier studies~\cite{ho2022VDM, ho2022imagen} explored the 3D U-Net and cascaded models for diffusion within pixel space. 
Recent solutions~\cite{chen2023videocrafter1, rombach2022high} employed latent diffusion models to efficiently manage the diffusion process within a compressed latent space. 
Following these studies, a variety of approaches~\cite{wang2023modelscope, blattmann2023align, lin2024ctrl, wang2023lavie, girdhar2023emu, wang2023videofactory, yu2023video, mei2023vidm, li2021keep} updated and improved this paradigm. 
Building on these advancements, subsequent methods further explored generating videos of higher quality and extended duration.
The Videocrafter approach~\cite{chen2024videocrafter2} pursued high-quality video generation through disentangling spatial and temporal learning and tuning spatial modules using high-quality images. 
In a similar way, commercial models such as Pika~\cite{pika} and GEN-2~\cite{gen2} demonstrated substantial improvements, showcasing videos with exceptional visual clarity. 
For longer video generation, Gen-L-Video~\cite{wang2023gen} aggregated short clips generated by base T2V models using temporal co-denoising to enhance continuity. 
Freenoise~\cite{qiu2023freenoise} extended pre-trained T2V models through rescheduling noise for longer-duration video inference. 
StreamingT2V~\cite{henschel2024streamingt2v} enhanced long-term content consistency by integrating short-term and long-term memory blocks.

The rapid development of T2V models poses a growing demand for quality evaluation protocols. Unfortunately, existing protocols primarily focus on temporal consistency and content continuity, yet largely ignore temporal dynamics. 
This hinders the exploitation of video content vividness and the honesty of video content to text prompts.

\subsection{Evaluation Protocol}
Early evaluation protocols~\cite{su2009ucf} primarily relied on class labels to evaluate the performance of T2V generation models.
For example, they commonly used video clips from the UCF-101 dataset and human-annotated video captions from the MSR-VTT~\cite{xu2016msr} dataset as the evaluation data.
For a more specific assessment, FETV~\cite{liu2024fetv} assigned fine-grained category labels to prompts and calculated the CLIP-SIM score for each category.

However, conventional quality assessment metrics such as Inception Score (IS)~\cite{salimans2016improved}, Fréchet Inception Distance (FID)~\cite{heusel2017gans}, Frechet Video Distance (FVD)~\cite{unterthiner2019fvd}, and CLIP-SIM typically operate on a single dimension while can not provide a comprehensive evaluation.
When addressing the limitation, EvalCrafter~\cite{radford2021learning} expanded both the prompt scale and the number of evaluation metrics so that the text-video alignment degree and the quality of generated videos can be better evaluated. 
Additionally, VBench~\cite{huang2023vbench} proposed a multi-dimensional, multi-category evaluation suite that not only considered the diversity of prompts but also encompassed a variety of assessment metrics.

Despite of the evolution of evaluation metrics, we argue an essential characteristic of video, $i.e.$, dynamics, remains ignored. In this study, we introduce the dynamics dimension to evaluate T2V generation models, as well as enhance the completeness of existing metrics. 

\section{Dynamics Evaluation Protocol}
\label{sec:dynamics_evaluation_protocol}

\begin{wraptable}{r}{0.55\textwidth}
\vspace{-1.4cm}
\centering
\footnotesize
\caption{Symbol Definitions.}
\label{tab:symbol}
\begin{tabular}{ll}
\toprule
Symbol & Definition                                           \\
\midrule
$\mathbf{D}_{<name>}$      & <$name$>-type dynamic metric of T2V models. \\
$\mathcal{T}$      & Text prompt benchmark of our DEVIL.                          \\
$M$ & The number of text prompts $\mathcal{T}$. \\
$T^i$     & $i$-th text prompt in $\mathcal{T}$, where $i\in \{1,\cdots, M\}$. \\
$G^i$      & Dynamic grade of text prompt $T^i$.\\
$S^i$      & Dynamics score of video generated by $T^i$.  \\
$f_j$      & The $j$-th video frame. \\
$F_j$      & Feature of $j$-th video frame. \\
$N$ & The number of video frames. \\
\bottomrule
\end{tabular}
\end{wraptable}

In this section, we first provide an overview of the proposed DEVIL protocol in Section~\ref{sec:overview} and then introduce the dynamics metrics proposed within DEVIL in Section~\ref{sec:dynamic_metrics}
Finally, we detail the prompt benchmark(Section~\ref{sec:benchmark}) and dynamics scores(Section~\ref{sec:dynamic_scores} and ~\ref{sec:aligned_dynamics_score}) constructed to evaluate the dynamics metrics of T2V generation models.

\subsection{Overview}
\label{sec:overview}
\begin{figure}
  \begin{center}
    \includegraphics[width=1.0\linewidth]{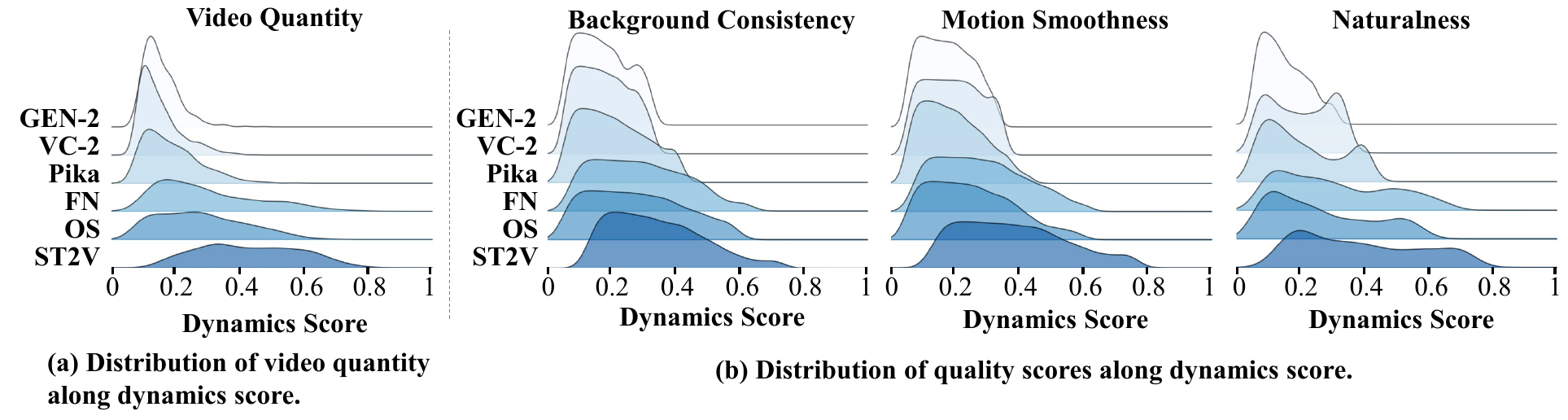}
  \end{center}
  
 \caption{Distributions of video quantity and quality scores along the dynamics score for various video generation models including: GEN-2~\cite{gen2}, Pika~\cite{pika}, VideoCrafter2(VC-2)~\cite{chen2024videocrafter2}, Open-Sora(OS)~\cite{opensora}, StreamingT2V~\cite{henschel2024streamingt2v} and FreeNoise-Lavie(FN)~\cite{qiu2023freenoise}.
 Subplot (a) shows video quantity distribution. Subplots (b) display the distribution of quality score of generated videos in terms of Background Consistency, Motion Smoothness, and Naturalness, respectively. 
All videos are generated based on our text prompt benchmark.}
  \label{fig:sta_dynamics}
\end{figure}

Fig.~\ref{fig:flowchat} shows the evaluation workflow of the DEVIL protocol. 
We aim to calculate the three dynamics metrics, \textit{dynamics range} (\( \mathbf{D}_{range} \)), \textit{dynamics controllability} (\( \mathbf{D}_{control} \)), and \textit{dynamics-based quality} (\( \mathbf{D}_{quality} \)) for each T2V model.
To achieve this, we establish a text prompts benchmark $\mathcal{T}=\{(T^i, G^i)\}_{i=1}^M$, where each prompt $T^i$ has a dynamic grade $G^i$, classified by GPT-4~\cite{openai2023chatgpt}, followed by further manual refinement.
$M$ is the number of prompts, for which we collect around 800 text prompts for our benchmark.
Subsequently, we generate videos using $\mathcal{T}$, and assess the dynamics of each generated video using an overall dynamics score $S$.
To calculate $S$, we define a series of dynamics scores at different temporal granularities, including inter-frame, inter-segment, and video levels, to reveal the video characteristics at multiple temporal levels as shown in Table~\ref{tab:dynamic_scores}. 
These scores are combined to obtain $S$ using weights derived from fitting human ratings.
Subsequently, the dynamics scores of all generated videos are utilized to calculate the three dynamics metrics, which represent the overall performance of T2V models.
In simplification, we provide the symbol definitions in Table~\ref{tab:symbol}.

\subsection{Dynamics Metrics}
\label{sec:dynamic_metrics}

We introduce three key metrics, \textit{dynamics range} (\( \mathbf{D}_{range} \)), \textit{dynamics controllability} (\( \mathbf{D}_{control} \)), and \textit{dynamics-based quality} (\( \mathbf{D}_{quality} \)), to evaluate T2V models from the perspective of dynamics. 
Each of these metrics evaluates the overall benchmark (described in Section~\ref{sec:benchmark}), which is calculated using the per-video dynamics scores (detailed in Sections~\ref{sec:dynamic_metrics} and ~\ref{sec:aligned_dynamics_score}).

\textbf{(i) Dynamics Range} demonstrates the model's versatility in handling both subtle and dramatic changes.
An ideal T2V generation model is expected to display a large dynamics range, reflecting various temporal variations described in text prompts.

In detail, we determine the dynamics range metric $\mathbf{D}_{range}$ by identifying the extremes of the dynamic scores over the benchmark, while excluding the top and bottom 1\% scores to mitigate the influence of outliers. 
This is formulated as
\begin{equation}
    \mathbf{D}_{range} = \mathbf{Q}_{0.99} - \mathbf{Q}_{0.01},
\end{equation}
where $\mathbf{Q}_{0.99}$ and $\mathbf{Q}_{0.01}$ denote the $99$-th and $1$-st percentile values of the dynamics scores for videos generated with our proposed text prompt benchmark, respectively. This metric reflects a realistic spread of dynamics, excluding atypical extremes.

\textbf{(ii) Dynamics Controllabiliy} assesses the ability of T2V models to manipulate video dynamics with text prompts. 
Objectively, it is challenging to obtain an exact correspondence between text prompts and videos.
Therefore, we adopt a ranking consistency-based methodology to derive a Dynamics Controllability metric $\mathbf{D}_{control}$.

Specifically, for two text prompts $T^i$ and $T^j$ in benchmark $\mathcal{T}=\{(T^i, G^i)\}$, their corresponding generated videos have dynamics scores $S^i$ and $S^j$ (the dynamics scores are detailed in Section~\ref{sec:aligned_dynamics_score}). 
Provided that the dynamics grades are ranked $G^i > G^j$, the dynamics scores should consequently be consistently ranked $S^i > S^j$. 
Accordingly, we calculate $\mathbf{D}_{control}$ as follows:
\begin{equation}
    \mathbf{D}_{control} = \frac{1}{M}\sum_{i=1}^{M}{\frac{1}{{M}-{M^i}}\sum_{j:G^j\neq G^i}{\mathbb{I}\big((S^{i}-S^{j})(G^{i}-G^{j})\big) }},
\end{equation}
\begin{wraptable}{r}{0.5\textwidth}
\centering
\footnotesize
\caption{Correlation between the overall dynamic score and the existing quality metrics, including 
Naturalness (Nat), Visual Quality~\cite{wu2023dover} (VQ), Motion Smoothness (MS)~\cite{huang2023vbench}, Subject Consistency (SC)~\cite{huang2023vbench} and Background Consistency (BC)~\cite{huang2023vbench}.
``PC'' denotes Pearson’s correlation, and ``KC'' denotes Kendall’s correlation.
}
\label{tab:corr-quality-dynamics}
\resizebox{0.48\textwidth}{!}{
\begin{tabular}{r|cc}
\toprule
Evaluation Metrics                    & PC & KC  \\ 
\midrule
Naturalness (Nat)                    &  -51.8    & -44.2     \\ 
Visual Quality (VQ)           &  -24.8     &   -18.6     \\
Motion Smoothness (MS)         & -64.0    & -54.6      \\
Subject Consistency (SC)       &  -88.9     & -74.9    \\ 
Background Consistency (BC)    & -79.4     & -61.4 \\
\bottomrule
\end{tabular}
\vspace{-1.3cm}
}
\end{wraptable}
where $M$ is the number of all text prompts and $M^i$ denotes the set of prompts with a dynamics grade of $G^i$. 
$\mathbb{I}(\cdot)$ denotes the indicator function.

\textbf{(iii) Dynamics-based Quality.}
Existing evaluations of generated visual quality do not account for the dynamics of the videos.
Previous studies~\cite{huang2023vbench, liu2023evalcrafter} have shown that videos with higher dynamics tend to receive lower quality scores.
In Table~\ref{tab:corr-quality-dynamics}, we calculate the correlation between the overall dynamics score of each generated video (as detailed in Section~\ref{sec:aligned_dynamics_score}) and its quality metrics. 
In detail, quality metrics such as Naturalness (Nat., elaborated in Section~\ref{sec:paper_naturalness}) , Motion Smoothness (MS)~\cite{huang2023vbench}, Subject Consistency (SC)~\cite{huang2023vbench}, and Background Consistency (BC)~\cite{huang2023vbench} exhibit a strong negative correlation with dynamics. 
This indicates that T2V models tend to generate low-dynamics videos for most text prompts to ``cheat'' to achieve higher scores on these metrics, as shown in Fig.~\ref{fig:sta_dynamics}.

To address this, we propose the Dynamics-based Quality metric $\textbf{D}_{quality}$, assessing generated visual quality considering dynamics.
For each video, we synthesize a composite quality score by averaging the scores of the identified quality metrics correlated with dynamics ($i.e.$, Nat, MS, SC, and BC).
We then divide the entire range of dynamics score into $L=12$ equal intervals and assign videos to their corresponding intervals based on their dynamics scores. 
Within each interval $l$, we calculate the average of the composite quality scores, denoted as $C_l$. 
Ultimately, the dynamic quality is defined as the overall average of these interval averages:
\begin{equation}
    \mathbf{D}_{quality} = \frac{1}{L} \sum_{l=1}^LC_l
\end{equation}
Except for dynamics-based quality on the entire range of dynamics score, we also evaluate dynamics-based quality at dynamics levels of high, medium, and low by modifying the range of intervals for a comprehensive evaluation (refer to Section~\ref{sec:exp-dynamics-metrics}).
Upon the dynamics-based quality, to have a high score, the generated videos should spread all dynamics intervals, which implies a large dynamics range. 
Additionally, for detailed results that integrate the dynamics score with individual metrics, please refer to Appendix~
\ref{app:detail_dynamics-based-qulity}.

\noindent{\textbf{Naturalness}.\label{sec:paper_naturalness}}
We propose Naturalness metric to evaluate the ability of T2V models to generate realistic videos.
In video generation, increased video dynamics often lead to unnatural phenomena, like a cat with an extra leg or water flowing uphill. 
Existing metrics focus on visual effects, ignoring video naturalness. 
However, a model's ability to generate natural videos reflects its real-world simulating ability. 
To assess this, we use the multi-modal model, Gemini 1.5 Pro~\cite{gemini2024}, to grade each video's naturalness into five levels
\footnote{Please refer to Appendix~\ref{sec:naturalness} for more details.}
: ``Almost Real'', ``Slightly Unrealistic'', ``Moderately Unrealistic'', ``Noticeably Unrealistic,'' and ``Completely Fictitious''. 
The overall naturalness is the average score of all videos. 
Experiments (see Table~\ref{tab:ablation}) show a high correlation between our scores and human ratings, validating the metric's effectiveness.

\subsection{Text Prompt Benchmark}
\label{sec:benchmark}

To evaluate the proposed dynamics metrics, we need a benchmark consisting of text prompts that fully represent multiple dynamic grades.
Existing benchmarks~\cite{huang2023vbench,liu2023evalcrafter} can not explicitly reflect various dynamics.
To this end, we establish a new benchmark.
Let $\mathcal{T}=\{(T^i, G^i)\}_{i=1}^N$ denote the benchmark, where each text prompt $T^i$ is assigned a dynamic grade $G^i$. Here, $G^i\in\{1,2,3,4,5\}$ that is categorized into a coarse range.
%
The dynamic grades are defined based on the level of dynamics described in the text prompts:
"$1$" represents \textbf{Static video}, where the video content is nearly stationary;
"$2$" represents \textbf{Low dynamics}, indicating slow and slight changes in the video content;
"$3$" represents \textbf{Medium dynamics}, characterized by noticeable activity and changes but relatively smooth overall;
"$4$" represents \textbf{High dynamics}, with fast actions and changes;
and "$5$" represents \textbf{Very high dynamics}, indicating extremely rapid and frequent changes in the video content. 
\begin{wrapfigure}{r}{0.45\textwidth} 
  \includegraphics[width=\linewidth]{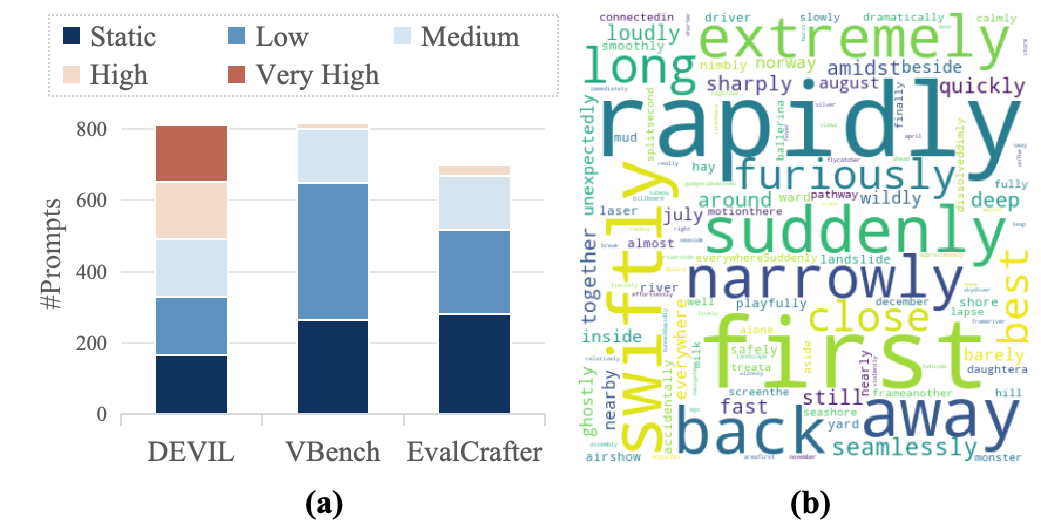}
  \caption{Dynamics distribution and Word cloud of text prompts from DEVIL, Vbench~\cite{huang2023vbench}, and EvalCrafter~\cite{liu2023evalcrafter}.}
  \label{fig:dynamic-distribution}
  \vspace{-0.5cm}
\end{wrapfigure}
In the coarse categorization step, we collect about 50,000 text prompts from existing benchmarks, including VidProm~\cite{wang2024vidprom}, WebVid~\cite{bain2021webvid}, MSR-VTT~\cite{xu2016msr}, and Didemo~\cite{hendricks18didemo}. 
The initial dynamic grades for each text prompt $T^i$ are assigned by GPT-4.
Then we recruit six human annotators for refinement for the post-processing step.
Finally, we sample 800 text prompts evenly across different dynamic grades to ensure a uniform distribution.

Fig.~\ref{fig:dynamic-distribution}(b) shows the statistics of the DEVIL benchmark, which contains approximately 800 text prompts, and each dynamics grade includes 19 object categories and 4 scene categories. 
For comparison, we further assign dynamic grades to the text prompts from existing benchmarks~\cite{huang2023vbench, liu2023evalcrafter} following the same procedure.
As shown in Fig.~\ref{fig:dynamic-distribution}(a), these benchmarks are heavily skewed towards lower dynamic content, while our benchmark demonstrates a more balanced distribution across all dynamic grades.
Unless otherwise specified, all experiments in this paper are conducted on the DEVIL benchmark.

\begin{table}[]
    \centering
    \footnotesize
    \caption{Formulations of dynamics scores at different temporal granularities.}
    \label{tab:dynamic_scores}
    \begin{tabular}{cll}
     \toprule
    Granularity        & Dynamics scores        & Formulation \\
    \midrule
    \multirow{4}{*}{Inter-frame}   & Optical Flow Strength    & $S_{ofs} =  \frac{1}{N-1} \sum_{i=1}^{N-1} \text{FLOW}(f_i)$ \\ \cmidrule{2-3}
                             & Structural Dynamics Score    & $S_{sd} = 1-\frac{1}{N-1}\sum_{i=1}^{N-1} \text{SSIM}(f_i, f_{i+1})$         \\ \cmidrule{2-3}
                             & Perceptual Dynamics Score    & $S_{pd} = \frac{1}{N-1}\sum_{i=1}^{N-1} \text{PHASHD}(f_i, f_{i+1})$         \\ \midrule
    \multirow{3}{*}{Inter-segment} & Patch-level Aperiodicity       & $S_{pa} = 1 - \frac{1}{HW} \sum_{h,w} \mathbf{ACF}(\{F_{i,h,w}\}_{i=1}^N)$  \\ \cmidrule{2-3}
                             & Global Aperiodicity        & $S_{ga}=1-\frac{1}{\lfloor rN \rfloor}\sum_{i=1}^{\lfloor rN \rfloor}\sum_{j\neq i}\mathbf{SIM}(F_i^r,F_j^r)$         \\ \midrule
    \multirow{3}{*}{Video}   & Temporal Entropy  & $S_{te} =  \mathbf{H}(f_1, f_2, \cdots, f_N|f_1)$         \\ \cmidrule{2-3}
                             & Temporal Semantic Diversity    & $S_{tsd} = \frac{1}{N}\sum_{i=1}^N \Vert F_i - \bar{F} \Vert^2$         \\
    \bottomrule
    \end{tabular}
\end{table}

\subsection{Dynamics Scores for Generated Videos}
\label{sec:dynamic_scores}

\begin{wrapfigure}{r}{0.45\textwidth} 
\vspace{-1cm}
  \includegraphics[width=\linewidth]{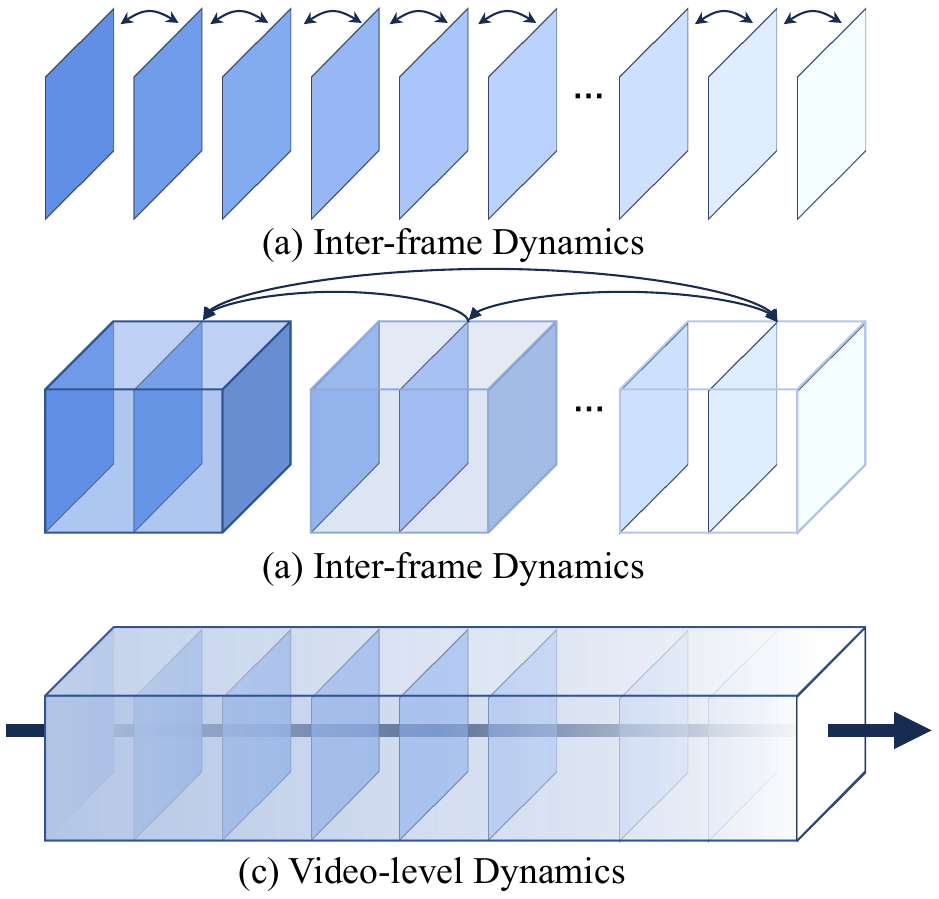}
  \caption{Video dynamics at different temporal granularities: (a) Inter-frame Dynamics, (b) Inter-segment Dynamics, and (c) Video-level Dynamics.}
  \label{fig:dynamic-granularity}
  \vspace{-3cm}
\end{wrapfigure}

To evaluate the proposed dynamics metrics, we generate videos using the text prompts from $\mathcal{T}=\{(T^i, P^i)\}_{i=1}^N$ and assess the dynamics of each generated video using a set of dynamics scores designed at different temporal granularities.
Specifically, we evaluate dynamics at three levels: inter-frame, inter-segment, and the entire video. 
By combining these evaluations, we derive an overall dynamics score.
For simplicity, we omit the superscripts from the dynamics scores in this section.

\textbf{(i) Inter-frame Dynamics Scores.} 
These scores describe variations between successive frames and are further divided into: \textit{optical flow strength}, \textit{structural dynamics}, and \textit{perceptual dynamics}. %

\textit{Optical flow strength.}
%
We first employ RAFT~\cite{zhang2024raft} to estimate the optical flow for each video frame.
%
The mean optical flow magnitudes of each frame are averaged to calculate the optical flow strength of this frame.
Averaging the optical flow strength values of all video frames, we have the optical flow strength $S_{ofs}$ of the video, as
\begin{equation}
    S_{ofs} = \frac{1}{N-1} \sum_{i=1}^{N-1} \text{FLOW}(f_i),
\end{equation}
where $\text{FLOW}$ calculate the mean optical flow strength values of frame $f_i$.

\noindent{\textit{Structural dynamics score}.}
%
While optical flow excels in capturing motion, it is less effective when detecting structural dynamics such as lighting conditions.
To capture such information, we calculate the average structural similarity index metric (SSIM)~\cite{ssim} between consecutive frames from all frame pairs to quantify inter-frame structural variations of the video, as
\begin{equation}
    S_{sd} = 1-\frac{1}{N-1} \sum_{i=1}^{N-1} \text{SSIM}(f_i, f_{i+1}).
\end{equation}
{\noindent{\textit{Perceptual dynamics}.} 
The human visual system is sensitive to changes in low-frequency regions of video frames. 
To reflect this characteristic, we introduce a perceptual dynamics score that measures the difference between the perceptual hashes~\cite{venkatesan2000imagehash} of consecutive frames. 
The perceptual distance $D_{pa}$ is defined as the mean perceptual hash distance of all frame pairs, as
\begin{equation}
    S_{pd} = \frac{1}{N-1} \sum_{i=1}^{N-1} \text{PHASHD}(f_i,f_{i+1}) ,
\end{equation}
}
where $\text{PHASHD}(f_i,f_{i+1})$ denotes the Hamming distance~\cite{hamming1950error} between the perceptual hash of $f_i$ and $f_{i+1}$.

\textbf{(ii) Inter-segment Dynamics Scores.}
These scores refer to the changes between video segments, each containing multiple frames. They capture the patterns of video content changes and are further categorized into \textit{patch-level aperiodicity} and \textit{global aperiodicity}, which measure the dynamics between video segments. %

\noindent{\textit{Patch-level aperiodicity}.}
We first calculate inter-segment dynamics at the patch level using the auto-correlation factor~\cite{box2015time}($\mathbf{ACF}$), to evaluate the scene and temporal pattern dynamics.
The auto-correlation factor measures the feature similarity of a time series, revealing periodicity and changing trends of features. 
Given features at position $(h, w)$ across $N$ frames, $\{F_{i, h, w}\}_{i=1}^{N}$, the auto-correlation factor of the features is defined as
\begin{equation}
    \mathbf{ACF}(\{F_{i,h,w}\}_{i=1}^N) = \frac{1}{N-K_0}\sum_{k=K_0}^N\sum_{i=1}^k\frac{1}{k}\mathbf{SIM}(F_{i,h,w}, F_{N-k+i,h,w}),
\end{equation}
where $K_0$ is the minimal segment length.
$\mathbf{SIM}$ represents the cosine similarity between two feature vectors.
It is empirically set to $\lfloor N/8\rfloor$ because most generated videos have more than 8 frames. 
$H$ and $W$ are the height and width of the feature map, respectively.
With auto-correlation factors of all patches, we define the patch-level aperiodicity of the video, as
\begin{equation}
    S_{pa} = 1 - \frac{1}{HW} \sum_{h,w} \mathbf{ACF}(\{F_{i,h,w}\}_{i=1}^N\}).
\end{equation}

\textit{Global aperiodicity.} 
In addition to patch-level dynamics, we employ a global aperiodicity score to measure the diversity of patterns between video segments. 
Specifically, we divide the video into segments. Each segment has a length $rN$, where $r$ is a proportion factor, empirically set to 0.25. We use ViCLIP~\cite{wang2023internvid} to extract the spatial-temporal features for each segment. The features are denoted as $\{F_i^r\}_{i=1}^{\lfloor rN \rfloor}$. 
We then calculate the similarity of these features to assess the variation in spatial-temporal patterns across segments, as
\begin{equation}
    S_{ga}=1-\frac{1}{\lfloor rN \rfloor}\sum_{i=1}^{\lfloor rN \rfloor}\sum_{j\neq i}\mathbf{SIM}(F_i^r,F_j^r).
\end{equation}

\textbf{(iii) Video-level Dynamics Scores.}
These scores encompass the overall content diversity and the frequency of changes throughout the video. 
The dynamics scores at video-level are defined by \textit{temporal entropy} and \textit{temporal semantic dynamics}.

%
\noindent{\textit{Temporal entropy}.
To evaluate the dynamics at the video level, we first measure the temporal information of each video. 
The temporal information $\mathbf{H}$ is defined as the conditional entropy of the entire video sequence given the first frame
\begin{equation}
    S_{te} =  \mathbf{H}(f_1, f_2, \cdots, f_N|f_1).
\end{equation}
%
To estimate the conditional entropy $S_{te}$, we employ the video encoding toolbox FFmpeg~\cite{ffmpeg}.

\textit{Temporal Semantic Dynamics}.
Beyond low-level dynamics, we further introduce a semantic diversity score to assess high-level dynamics across the whole video.
The semantic diversity score $S_{tsd}$ is computed to reflect semantic-level dynamics and is defined as the variance of DINO~\cite{DINO2021} features $\{F_i\}_{i=1}^N$ of each frame, as
\begin{equation}
    S_{tsd} = \frac{1}{N}\sum_{i=1}^N \Vert F_i - \bar{F} \Vert^2, \ 
\end{equation}
where $\bar{F}=\frac{1}{N}\sum_{i=1}^N{F_i}$ denotes the mean feature vector of all frames. 

\subsection{Overall Dynamics Score}
\label{sec:aligned_dynamics_score}

To establish a reliable and robust assessment, we integrate dynamics scores into one with a human alignment procedure, Fig.~\ref{fig:flowchat}, to refine the empirically defined dynamics score.
It utilizes human ratings to provide ground-truth, based on which we fit a linear regression model at each temporal granularity, as 
\begin{align}
S_{f} &= \mathbf{Linear}_{\theta_{f}}(D_{ofs}, D_{sd}, D_{pd}), \\
S_{s} &= \mathbf{Linear}_{\theta_{s}}(D_{pa}, D_{ga}), \\
S_{v} &= \mathbf{Linear}_{\theta_{v}}(D_{te}, D_{tsd}),
\end{align}
where $\theta_{f}, \theta_{s}, \theta_{v}$ respectively denote the model parameters of linear regression at each granularity.
The overall dynamics score of the video is then defined as the average of aligned dynamics scores from all three levels, as
\begin{equation}
\label{eq:aligned_dynimics}
    S = \frac{1}{3}(S_{f} + S_{s} + S_{v}).
\end{equation}
Through this learnable human alignment procedure, the empirically defined dynamics scores are more consistent with human perception, as validated in Sec.~\ref{sec:ablation}.

\begin{wraptable}{r}{0.45\textwidth}
\vspace{-1.1cm}
\centering
\footnotesize
\caption{Human alignment by correlation between dynamics scores and human ratings on the proposed DEVIL benchmark. Video generation is based on text prompts in DEVIL. 
``PC'' denotes Pearson's correlation, ``KC'' Kendall's correlation, and ``WR'' the win ratio.}
\resizebox{0.43\textwidth}{!}{
\begin{tabular}{ll|ccc}
\toprule
\multicolumn{2}{l}{Scores}                                                         & PC $\uparrow $ & KC $\uparrow $ & WR $\uparrow $ \\ \midrule
\multicolumn{1}{l}{\multirow{4}{*}{Inter-frame}}   & $S_{ofs}$              & 93.1        & 89.9       &  79.2           \\
\multicolumn{1}{c}{}                                     & $S_{sd}$                   & 91.7        & 88.0       &  78.1           \\
\multicolumn{1}{c}{}                                     & $S_{pd}$                    & 96.4        & 93.2       & 86.1  \\ 
\multicolumn{1}{c}{}                                     & $S_{f}$                    & \textbf{96.5}        & \textbf{93.5}       & \textbf{86.5}              \\ 
\midrule
\multicolumn{1}{l}{\multirow{3}{*}{Inter-segment}} & $S_{pa}$                          & 95.1        &  94.3      & 87.0            \\
\multicolumn{1}{c}{}                                     & $S_{g}$                    & 94.6        &  93.0      & 85.6            \\
\multicolumn{1}{c}{}                                     & $S_{s}$           & \textbf{95.8}        &  \textbf{94.8}      & \textbf{87.7}            \\ 

\midrule
\multicolumn{1}{l}{\multirow{3}{*}{Video level}}         & $S_{te}$     & 96.4        & 93.7      & 83.5            \\
\multicolumn{1}{c}{}                                     & $S_{tsd}$     & 97.7        & 96.4       & 90.5            \\   
\multicolumn{1}{c}{}                                     & $S_{v}$      & \textbf{98.0}        & \textbf{97.2}       & \textbf{91.4}            \\ 
\midrule
\midrule
\multicolumn{2}{l|}{Naturalness}    & 79.0 & 75.5 & 52.4 \\
\bottomrule
\end{tabular}
}
\label{tab:ablation}
\vspace{-2cm}
\end{wraptable}

\section{Experiments}

\subsection{Human Alignment Assessment}
\label{sec:ablation}

To evaluate the plausibility of the proposed dynamics metrics and the naturalness metric, we conduct the following human alignment experiments.

\noindent{\textbf{Ground-truth Annotation}.}
We first generate videos using six state-of-the-art (SOTA) T2V models, including GEN-2~\cite{gen2}, Pika~\cite{pika}, VideoCrafter2~\cite{chen2024videocrafter2}, Open-Sora~\cite{opensora}, StreamingT2V~\cite{henschel2024streamingt2v} and FreeNoise-Lavie~\cite{qiu2023freenoise} and DEVIL text prompts.
For the generated videos, we collect human evaluated dynamics and naturalness as the ground-truth.
Six persons are recruited to assess each video's grade of dynamics under three temporal levels (Frame, Segment and Video).
For each dynamics metric, evaluators are required to rate the grade of dynamics from ``static'' to ``very high dynamics'' defined in Section~\ref{sec:benchmark}. 
To guide the annotation process, we provide specific prompts for each temporal level. \footnote{Please refer to Appendix ~\ref{sec:human-annotation} for details}.
The evaluation of the naturalness metric follows the same process, where a higher human assigned grade indicates a greater degree of naturalness. 

\noindent{\textbf{Evaluation of Scores}.}
We calculate dynamics grades and naturalness for generated videos on the proposed DEVIL benchmark. 
For dynamics metrics at multiple temporal levels, we integrate them using the linear regression model defined by Eq.~\ref{eq:aligned_dynimics}.
For each linear regression model, it takes the human evaluation results as ground-truths, trained upon 75\% of the randomly selected videos and tests on the remaining 25\% videos.
During testing, the human alignment performance is reflected by the correlation $e.g.$, Pearson and Kendall's correlation coefficients and win ratio, between predicted and human evaluated dynamics grades.
The win ratio involves comparing each video against others with different grades of dynamics. For instance, a video rated as ``high dynamics'' by evaluators should score lower in dynamics than any video rated as ``Very high dynamics'' but higher than those rated as ``static''. 

Table~\ref{tab:ablation} shows the assessment results of the six T2V generation models. It can be seen that the dynamics metrics and the naturalness metric exhibit a strong alignment with human evaluation.
The improved metrics ($S_f$, $S_s$, $S_v$ defined in Sec.~\ref{sec:aligned_dynamics_score}) further enhance the alignment with human evaluations.

\begin{table}[t]
  \centering
  \footnotesize
    \caption{Evaluation of T2V models on dynamics range (\(\mathbf{D}_{range}\)), dynamics controllability (\(\mathbf{D}_{control}\)), and dynamics quality (\(\mathbf{D}_{quality}\)) using our text prompt benchmark. All metrics are normalized with maximum values of 100\% and minimum values of 0\%, higher scores indicate better performance. Dynamics quality is also assessed at low (\(\mathbf{D}_{quality}^L\)), medium (\(\mathbf{D}_{quality}^M\)), and high (\(\mathbf{D}_{quality}^H\)) levels. }
\label{tab:metrics}
\begin{tabular}{l|c|c|c|ccc}
\toprule
T2V models      & $\mathbf{D}_{range}$ & $\mathbf{D}_{control}$ & $\mathbf{D}_{quality}$ & $\mathbf{D}_{quality}^{L}$ & $\mathbf{D}_{quality}^{M}$ & $\mathbf{D}_{quality}^{H}$ \\ \midrule
GEN-2~\cite{gen2}           &  30.8        &  \cellcolor{mycolor}82.5          & 43.6              & \cellcolor{mycolor}93.4             & 45.4               & 0.0                               \\
Pika~\cite{pika}            &  43.2        &  \cellcolor{mycolor!40}72.0          & 52.1              & 90.0             & 66.4               & 0.0                              \\
VideoCrafter2~\cite{chen2024videocrafter2}   &  34.1        &  57.0          & 43.6               & 89.1              & 41.7                & 0.0                              \\
OpenSora~\cite{opensora}        &  \cellcolor{mycolor!40}61.2        &  62.4          & \cellcolor{mycolor!70}63.7              & 84.4             & \cellcolor{mycolor!70}84.5              & \cellcolor{mycolor!40}22.2                              \\
StreamingT2V~\cite{henschel2024streamingt2v}    &  \cellcolor{mycolor!70}65.9        &  62.8          & \cellcolor{mycolor!20}60.8              & 61.1             & \cellcolor{mycolor!40}80.8              & \cellcolor{mycolor!70}40.6                               \\
FreeNoise-Lavie~\cite{qiu2023freenoise} & \cellcolor{mycolor}66.9         &  58.7          & \cellcolor{mycolor}66.3              & 65.9             & \cellcolor{mycolor}87.7              & \cellcolor{mycolor}45.5                               \\
Hotshot-XL~\cite{hotshot}      &  34.7        & 58.9           &  52.2             & \cellcolor{mycolor!70}92.8             & 63.9               & 0.0                              \\
Show-1~\cite{zhang2023show}          & 45.1         & \cellcolor{mycolor!70}73.9           & 57.7             & \cellcolor{mycolor!40}92.6             & \cellcolor{mycolor!20}80.3               & 0.0                             \\
ModelScope~\cite{wang2023modelscope}      & \cellcolor{mycolor!20}52.9         & 63.6           & \cellcolor{mycolor!40}62.6             & \cellcolor{mycolor!20}91.2             & 79.1              & \cellcolor{mycolor!20}17.5                              \\
ZeroScope~\cite{zeroscope}       & 26.4         & \cellcolor{mycolor!20}66.4           &  44.8             & 90.9             & 43.6               & 0.0                              \\ \bottomrule
\end{tabular}
\end{table}

\vspace{-0.2cm}
\subsection{Dynamic-Quality Bi-variate Analysis}
To investigate the relationship between video dynamics and quality, we calculated the correlation coefficients between various quality metrics and the overall dynamics score ($S$), as well as the distribution of video quality scores along $S$. 
As shown in Table~\ref{tab:corr}, Naturalness, Motion Smoothness, Subject Consistency, and Background Consistency all have Pearson correlation coefficients above 50\% with $S$, indicating the significant impact of dynamics on these metrics. 
Fig.~\ref{fig:sta_dynamics} shows the distribution of video quantity and quality scores along $S$. 
Most models, especially high-ranking ones like GEN-2~\cite{gen2}, Pika~\cite{pika}, and VideoCrafter2~\cite{chen2024videocrafter2}, generate videos concentrated in low dynamic regions. 
As dynamics increase, quality metrics significantly decline.
This suggests that models can improve benchmark quality scores by generating low-dynamic videos. 
In conclusion, video dynamics significantly impact quality evaluation, and quality metrics design should account for dynamics.

\subsection{Evaluation of Dynamics Metrics}
\label{sec:exp-dynamics-metrics}
\vspace{-0.2cm}

We evaluate the dynamics range $\mathbf{D}_{range}$, dynamics controllability
 $\mathbf{D}_{control}$ and dynamics quality $\mathbf{D}_{quality}$ of T2V models on our text prompt benchmark. All metrics are normalized with maximum values of 100\% and minimum values of 0\%.
To assess dynamics quality, we consider low, medium, and high levels, obtaining $\mathbf{D}_{quality}^L$, $\mathbf{D}_{quality}^M$, and $\mathbf{D}_{quality}^H$. The score ranges for these levels are [0, 33.3\%], [33.4\%, 66.7\%], and [66.8\%, 100\%] respectively, where higher scores indicate better performance.
The results are shown in Table~\ref{tab:metrics}. 
In addition to six models that are annotated, we also evaluate another five SOTA T2V models to provide a comprehensive comparison of the latest models.
It can be observed that the GEN-2~\cite{gen2} and Pika~\cite{pika} models achieve high dynamics alignment scores, but low dynamics range scores. 
This is because these methods generate videos with low dynamics. 
In contrast, the FreeNoise-Lavie~\cite{qiu2023freenoise} and StreamingT2V~\cite{henschel2024streamingt2v} achieve a high dynamics range but a low dynamics controllability score, indicating that it generates video dynamics misaligned with the text prompts.
\footnote{Please refer to Appendix~\ref{sec:corr-metric-dynamics} for details}.
 
\textbf{Origal Quality Metric v.s. Dynamics-based Quality Metric.} 
Fig.~\ref{fig:bar_quality} shows the comparison between the original quality metric and various dynamics-based quality metrics, including the overall dynamics-based quality metric ($\mathbf{D}_{quality}$) and metrics at low ($\mathbf{D}_{quality}^L$), medium ($\mathbf{D}_{quality}^M$), and high ($\mathbf{D}_{quality}^H$) dynamics levels. 
It shows that the original quality metric aligns closely with $\mathbf{D}_{quality}^L$, indicating that it primarily reflects quality in low dynamics scenarios. Moreover, T2V models typically lack the ability to generate high-dynamics videos, resulting in lower scores for $\mathbf{D}_{quality}^H$.

\begin{figure}
  \begin{center}
    \includegraphics[width=1.0\linewidth]{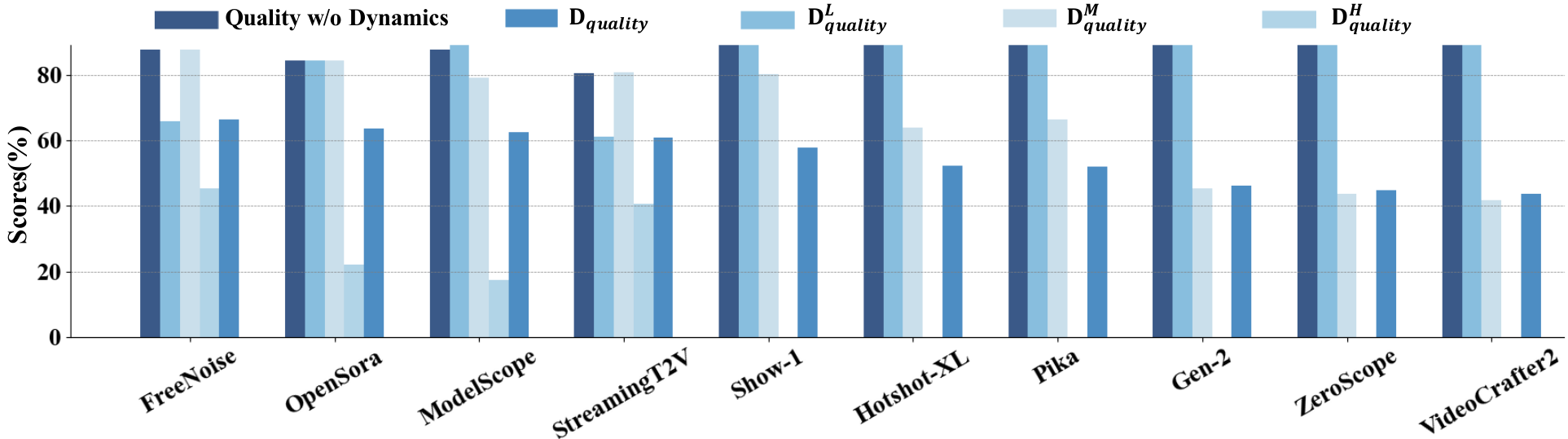}
  \end{center}
 \caption{Bar chart illustrating the original quality metric, overall dynamics-based quality metric ($\mathbf{D}_{quality}$) and dynamics-based quality metrics at low($\mathbf{D}_{quality}^L$), medium($\mathbf{D}_{quality}^M$) and high($\mathbf{D}_{quality}^H$) dynamics levels. (Best viewed in color)}
  \label{fig:bar_quality}
\end{figure}

\begin{wrapfigure}{r}{0.4\textwidth} 
  \vspace{-1.2cm}
  \includegraphics[width=\linewidth]{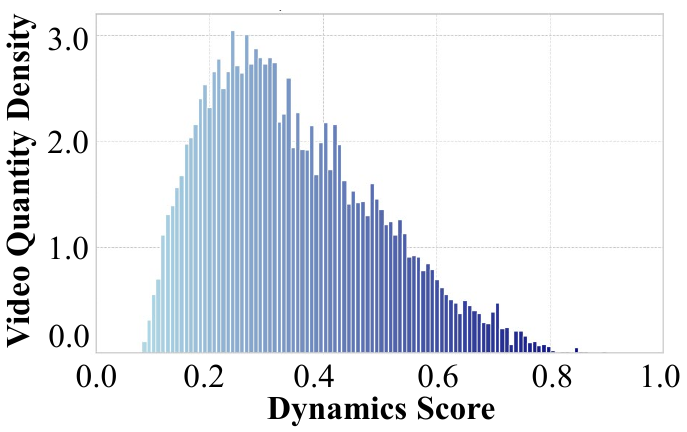}
  \caption{Video quantity density $w.r.t.$ dynamics score of the WebVid-2M dataset.}
  \label{fig:webvid-dynamic-distribution}
  \vspace{-0.5cm}
\end{wrapfigure}

\subsection{Insights from Video Dynamics Analysis}
\textbf{Existing datasets have biased dynamics distribution.}
The distribution of dynamics of the video datasets (such as WebVid2M~\cite{bain2021webvid}) is biased. The statistical result is shown in Fig.~\ref{fig:webvid-dynamic-distribution}. It can be seen that most of the videos have a small dynamics score ($\le$ 0.4).
The limited number of videos with high dynamics scores restricts the model's ability to generate dynamics-rich videos which are common in practical applications. \textit{Therefore, existing datasets should be expanded in terms of dynamics, and the proposed metrics can provide guidance for this expansion.}

\textbf{Existing datasets have biased text prompts on dynamics for training.}
We use the dynamics controllability metric to evaluate two popular datasets, $i.e.$, WebVid2M~\cite{bain2021webvid} and MSR-VTT~\cite{xu2016msr}, by using the ground-truth text prompts and videos.
Unfortunately, they respectively achieve dynamics controllability scores of 36.31\% and 52.98\%.
The poor performance indicates that the two datasets can not provide sufficient information/guidance while training the video generation models.
\textit{To train better video generation models, the text prompts of these datasets requires to be elaborated on aspects of dynamics.}

\textbf{Existing T2V methods have limited real-world simulation ability.}
As shown in Fig.~\ref{fig:sta_dynamics}, we performed a statistical analysis of video quantity distribution, visual quality, motion smoothness, and naturalness metric scores for SOTA methods based on the distribution of dynamics score.
When the dynamics score is small, the videos generated by these SOTA models have high scores under the aforementioned four metrics. 
As the dynamics score increases, these scores (especially the naturalness) significantly decrease. 
This might be caused by the fact that these models primarily focus on optimizing the generation of simple and slow-motion content, while dynamics are totally ignored in the evaluation metrics.
\textit{Therefore, T2V models should be optimized on large range of dynamics to truly reflect real-world simulation.}

\vspace{-0.2cm}
\section{Conclusion}
\vspace{-0.1cm}
We proposed DEVIL, a comprehensive and constructive evaluation protocol for T2V generation models.
In the protocol, we defined a set of dynamics metrics corresponding to multiple temporal granularities, and a new benchmark of text prompts under multiple levels of dynamics.
Based on the distribution of dynamics scores over the benchmark, we assessed the generation capacity of T2V models, characterized by dynamic ranges and degree of T2V alignment.
Experiments show that DEVIL enjoys 90\% consistency with human evaluation results, demonstrating the potential to be a powerful tool for advancing T2V generation models.

\textbf{Limitations.}  At present, the grades of dynamics remain limited, which should be improved to more fine-grained grades. Furthermore, only a limited number of T2V models are evaluated using the proposed protocol. A more comprehensive evaluation of T2V models should be done in future work.

\textbf{Social impacts}. The positive impact can be that the proposed evaluation protocol may promote the development of T2V models. The negative impact can be a risk that advanced T2V models could be misused to create realistic but misleading video content, such as deepfakes.

\bibliographystyle{plainnat}  
\bibliography{egbib}

\newpage

\appendix

\part*{Appendix}
\addcontentsline{toc}{part}{Appendix}

\section{Dynamics Scores}

\label{sec:corr-metric-dynamics}

\begin{figure}[htbp]
  \begin{center}
    \includegraphics[width=1.0\linewidth]{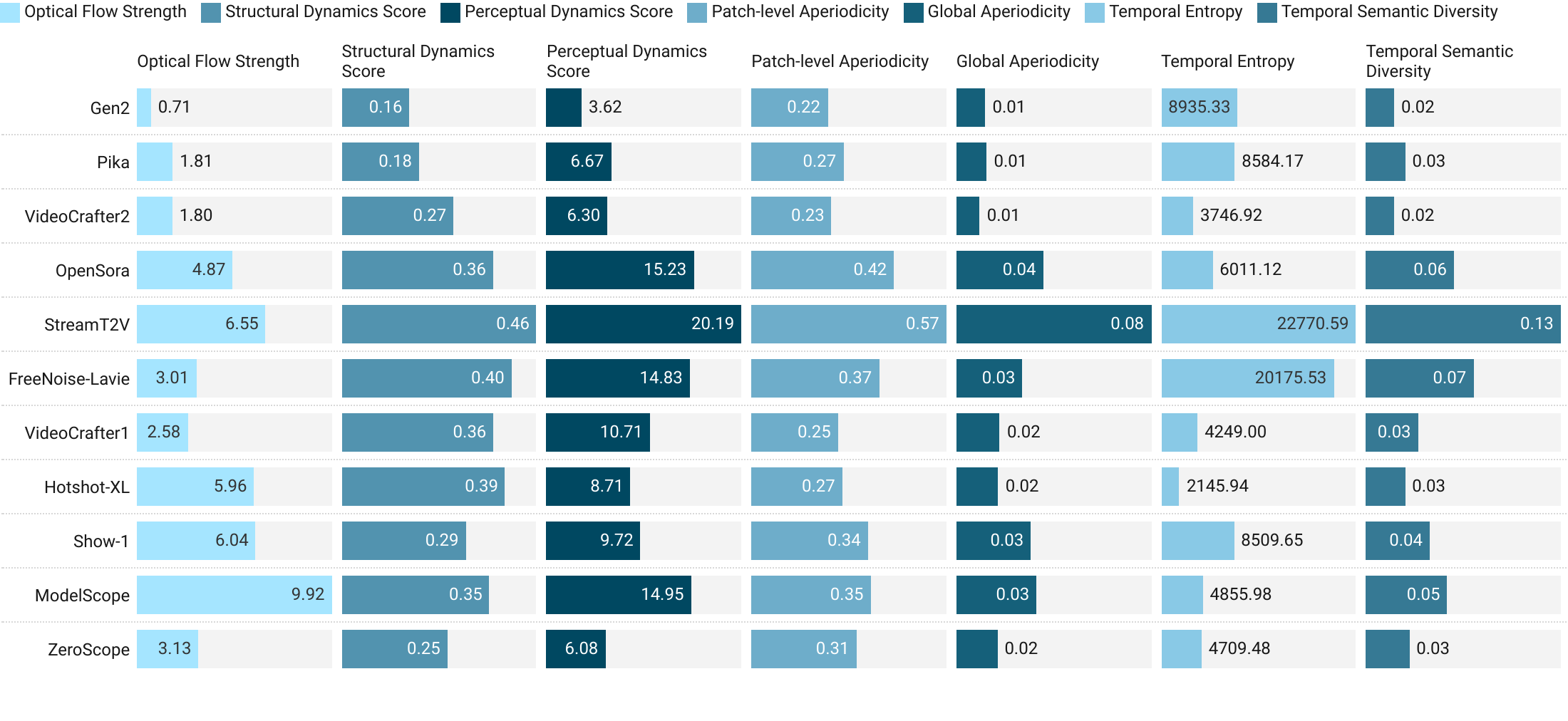}
  \end{center}
 \caption{Evauation of the state-of-the-art models using dynamics scores proposed in Section~\ref{sec:dynamic_metric}.}
 \label{fig:dynamic_score_details}
\end{figure}

For the dynamics scores proposed in Section~\ref{sec:dynamic_metric}, we present the detailed results of T2V models in Figure~\ref{fig:dynamic_score_details}.
It can be seen that ModelScope~\cite{wang2023modelscope} excels in generating rapid inter-frame motions, while StreamingT2V~\cite{henschel2024streamingt2v} performs exceptionally well across most dynamics score metrics.
StreamingT2V achieves high scores for the inter-segment dyanmics scores at video levels. 
This indicates that it has significant advantages in generating complex dynamic content. 
In contrast, GEN-2~\cite{gen2} and VideoCrafter2~\cite{chen2024videocrafter2} perform poorly on several metrics, highlighting their deficiencies in dynamics.
%

\section{Correlation Between Existing Metrics and Dynamics}

In Section~\ref{sec:dynamic_metrics}, to identify the relevance between existing metrics with the dynamics metrics, we provide a bi-variate analysis strategy.
Based on bi-variate analysis, we provide detailed correlation results for the models.
In Table~\ref{tab:corr}, the Pearson correlation coefficients between the dynamics scores and existing metrics, including aesthetic score, technical score, visual quality, motion smoothness, subject consistency, background consistency, and naturalness, are detailed.

The results indicate a clear trade-off between video dynamics and various existing metrics in T2V models.
As dynamic complexity increases, there tends to be a decline in motion smoothness, subject consistency, background consistency, and naturalness.
The aesthetic, technical, and visual quality metrics show relatively low correlation, which can be attributed to the fact that these metrics evaluate video frames independently, ignoring temporal relationships between frames.

\begin{table}[t]
\centering
\caption{Pearson correlation coefficient between the dynamics metrics and the existing metrics including aesthetic score~\cite{wu2023dover}, technical score~\cite{wu2023dover} visual quality~\cite{wu2023dover}, motion smoothness~\cite{huang2023vbench}, subject consistency~\cite{huang2023vbench} and background consistency~\cite{huang2023vbench} and our naturalness.}
\label{tab:corr}
\resizebox{\textwidth}{!}{
\begin{tabular}{l|ccccccc}
\toprule
                & \begin{tabular}[c]{@{}c@{}}Aesthetic \\ Score \end{tabular} 
                & \begin{tabular}[c]{@{}c@{}}Technical \\ Score \end{tabular} 
                & \begin{tabular}[c]{@{}c@{}}Visual \\ Quality \end{tabular} 
                & \begin{tabular}[c]{@{}c@{}}Motion \\ Smoothness\end{tabular} 
                & \begin{tabular}[c]{@{}c@{}}Subject \\ Consistency\end{tabular} 
                & \begin{tabular}[c]{@{}c@{}}Background \\ Consistency\end{tabular} 
                & \begin{tabular}[c]{@{}c@{}}Naturalness\end{tabular} \\ \midrule
GEN-2~\cite{gen2}            & -0.19  & -0.09   &  -0.12    & -0.54         & -0.88                & -0.73                       &  -0.50       \\
Pika~\cite{pika}            & -0.40   & -0.20   & -0.28      & -0.65        & -0.88               & -0.78                       &  -0.47       \\
VideoCrafter2~\cite{chen2024videocrafter2}   & -0.25  & -0.20    & -0.24     & -0.59        & -0.87               & -0.76                       &  -0.36       \\
OpenSora~\cite{opensora}        & -0.20   & -0.27   & -0.26     & -0.70         & -0.90                &  -0.83                       &  -0.43       \\
StreamingT2V~\cite{henschel2024streamingt2v}    & -0.15  & -0.21    & -0.23    & -0.57        & -0.89               &  -0.81                      &  -0.36       \\
FreeNoise-Lavie~\cite{qiu2023freenoise} & -0.37  & -0.31   & -0.35     & -0.75        & -0.91               &  -0.86                      &  -0.48       \\ \midrule
Average         & -0.26  & -0.21   & -0.25     & -0.63        & -0.81               & -0.79                      & -0.43 \\ \bottomrule
\end{tabular}
}
\end{table}

\section{Detail of Dynamics-based Quality}
\label{app:detail_dynamics-based-qulity}

Let $S^{(i)}$ denote a score of generated video $i$. Existing metrics simply average the scores of all videos to obtain the metric score $S$ of the $T2V$ model:
\begin{equation}
    S = \frac{1}{|T|} \sum_{i=1}^{|T|} S^{(i)},
\end{equation}
where $|T|$ is the total number of generated videos.
Considering that some existing metrics show a considerable negative correlation with the video's dynamics score, they fail to prevent models from generating low-dynamic videos.

To address this issue, we enhance existing metrics by integrating human-aligned dynamics scores, preventing models from attaining high scores by producing low-dynamic videos. 
Specifically, we first equally divide the human-aligned dynamics score into $L=12$ intervals. We then calculate the mean scores $S_l$ at each interval $l$. The improved metric $S^*$ is defined as the average of $S_l$ across all intervals:
\begin{equation}
    S^* = \frac{1}{L} \sum_{l=1}^L S_l.
\end{equation}

Table ~\ref{tab:quality_each_metric}  presents the scores of various models across four quality metrics: Motion Smoothness, Naturalness, Subject Consistency, and Background Consistency. FreeNoise and StreamingT2V achieve high overall scores due to their strong performance across a wide dynamic range. In contrast, Gen-2 and Pika excel in the low dynamic range, but their inability to generate high dynamic videos results in lower overall scores.

\begin{table}[]
\caption{Integrating dynamics scores with quality metrics, including Motion Smoothness, Naturalness, Subject Consistency, and Background Consistency. The table details scores across multiple models, with metrics divided into Overall, Low, Mid, and High categories based on modified dynamic intervals to achieve a comprehensive evaluation.}
\centering
\label{tab:quality_each_metric}
\begin{tabular}{l|cccc|cccc}
\toprule
\multirow{2}{*}{T2V Model} & \multicolumn{4}{c|}{MotionSmoothness}    & \multicolumn{4}{c}{Naturalness}            \\ \cmidrule{2-9} 
                      & Overall     & Low     & Mid     & High   & Overall     & Low      & Mid      & High   \\ \midrule
FreeNoise~\cite{qiu2023freenoise}             & 71.7        & 71.7    & 95.4    & 47.9   & 57.1        & 54.8     & 73.9     & 42.5   \\
GEN-2~\cite{gen2}                 & 49.7        & 99.5    & 49.7    & 0.0    & 39.1        & 81.6     & 35.6     & 0.0    \\
OpenSora~\cite{opensora}              & 71.5        & 95.5    & 95.3    & 23.7   & 49.8        & 62.8     & 64.2     & 22.5   \\
Pika~\cite{pika}                  & 58.0        & 99.5    & 74.5    & 0.0    & 39.8        & 69.4     & 50.1     & 0.0    \\
StreamingT2V~\cite{henschel2024streamingt2v}          & 71.2        & 70.9    & 95.0    & 47.8   & 42.2        & 44.2     & 55.4     & 27.0   \\
VideoCrafter2~\cite{chen2024videocrafter2}         & 48.9        & 97.8    & 48.8    & 0.0    & 31.4        & 70.1     & 24.2     & 0.0    \\
HotShot-XL~\cite{hotshot}            & 47.0        & 83.7    & 57.3    & 0.0    & 54.4        & 95.7     & 67.4     & 0.0    \\
ModelScope~\cite{wang2023modelscope}            & 50.4        & 77.1    & 61.6    & 12.5   & 67.9        & 95.7     & 87.6     & 20.4   \\
Show-1~\cite{zhang2023show}                & 47.4        & 81.6    & 60.6    & 0.0    & 62.0        & 95.5     & 90.5     & 0.0    \\
ZeroScope~\cite{zeroscope}             & 38.3        & 75.0    & 40.0    & 0.0    & 46.5        & 95.3     & 44.2     & 0.0    \\ \bottomrule \toprule
\multirow{2}{*}{T2V Model} & \multicolumn{4}{c|}{Subject Consistency} & \multicolumn{4}{c}{Background Consistency} \\ \cmidrule{2-9} 
                      & Overall     & Low     & Mid     & High   & Overall     & Low      & Mid      & High   \\ \midrule
FreeNoise~\cite{qiu2023freenoise}             & 66.0        & 66.3    & 87.3    & 44.3   & 70.6        & 70.7     & 94.0     & 45.5   \\
GEN-2~\cite{gen2}                 & 47.7        & 95.2    & 47.8    & 0.0    & 48.6        & 97.3     & 48.5     & 0.0    \\
OpenSora~\cite{opensora}              & 62.7        & 84.5    & 84.2    & 19.3   & 70.7        & 94.6     & 94.4     & 22.2   \\
Pika~\cite{pika}                  & 54.6        & 94.6    & 69.1    & 0.0    & 56.1        & 96.5     & 71.8     & 0.0    \\
StreamingT2V~\cite{henschel2024streamingt2v}          & 61.2        & 60.8    & 81.6    & 41.3   & 68.6        & 68.3     & 91.2     & 40.6   \\
VideoCrafter2~\cite{chen2024videocrafter2}         & 45.5        & 90.8    & 45.6    & 0.0    & 48.7        & 97.5     & 48.4     & 0.0    \\
HotShot-XL~\cite{hotshot}            & 55.6        & 97.1    & 69.7    & 0.0    & 52.0        & 94.8     & 61.1     & 0.0    \\
ModelScope~\cite{wang2023modelscope}            & 70.1        & 97.1    & 91.6    & 21.6   & 62.0        & 94.8     & 75.5     & 17.5   \\
Show-1~\cite{zhang2023show}                & 62.8        & 98.2    & 90.1    & 0.0    & 58.5        & 95.3     & 80.2     & 0.0    \\
ZeroScope~\cite{zeroscope}             & 49.0        & 98.4    & 48.5    & 0.0    & 45.5        & 94.8     & 41.7     & 0.0    \\ \bottomrule
\end{tabular}
\end{table}

\section{Assigning Dynamics Grades to Text Prompts}

As described in Section~\ref{sec:benchmark}, we collect approximately 50,000 text prompts from existing benchmarks, including 19 object categories and 4 scene categories. 
Using GPT-4 coarse classification and human refinement, we construct the DEVIL prompt benchmark. 
The process of categorizing dynamics grades using GPT-4 is illustrated in Figure~\ref{fig:prompt-categorize-gpt4}. 
In specific, we instruct GPT-4 to perform classification on the rate of content change.
To enhance GPT-4's classification accuracy, we further provide detailed criteria and examples for each dynamics grade. 
In the post-processing step, we recruit six human annotators to refine the dynamics grades over three months. 
Finally, we sample about 800 text prompts at different dynamics grades to ensure a uniform distribution across the grades.

\section{Details of Naturalness}
\label{sec:naturalness}
We employed the advanced multi-modal large model, Gemini-1.5 Pro~\cite{gemini2024}, equipped with video understanding capabilities, to assess and classify the naturalness of video content. 
As shown in Fig.~\ref{fig:gemini-naturalness}, we demonstrate the process through which the model analyzes videos and assigns naturalness ratings. 
The figure details the five different levels used to evaluate video naturalness, ranging from ``Completely Fantastical" to ``Almost Realistic". 
Each level is defined based on how closely the video content aligns with the real world. 
Additionally, the figure includes two examples of video evaluations: the first video is rated as "Almost Realistic" due to its high conformity with reality, while the second video, due to minor distortions—such as the unrealistic number of legs on a dog—is rated as "Slightly Unrealistic". 
These examples validate the plausibility of the proposed naturalness metric.

\section{Human Annotation}
\label{sec:human-annotation}
To align human evaluations with automated metrics, we annotated a series of videos generated by SOTA T2V models.
We initiated the process by generating videos using prompts from the DEVIL benchmark with six advanced T2V models including GEN-2, Pika, VideoCrafter2, OpenSora, StreamingT2V, and FreeNoise-Lavie.
Subsequently, we developed a video annotation toolbox for evaluating the dynamics and naturalness of videos. 
As shown in Figure~\ref{fig:ann-interface}, the toolbox allows annotators to assess the dynamics of the videos across five grades, from almost static to very high dynamics, and the naturalness from almost real to completely unreal. 
To guarantee high-quality and consistent evaluations, we recruit six annotators who have undergraduate degrees and provided them with detailed training.

\begin{figure}
  \begin{center}
    \includegraphics[width=1.0\linewidth]{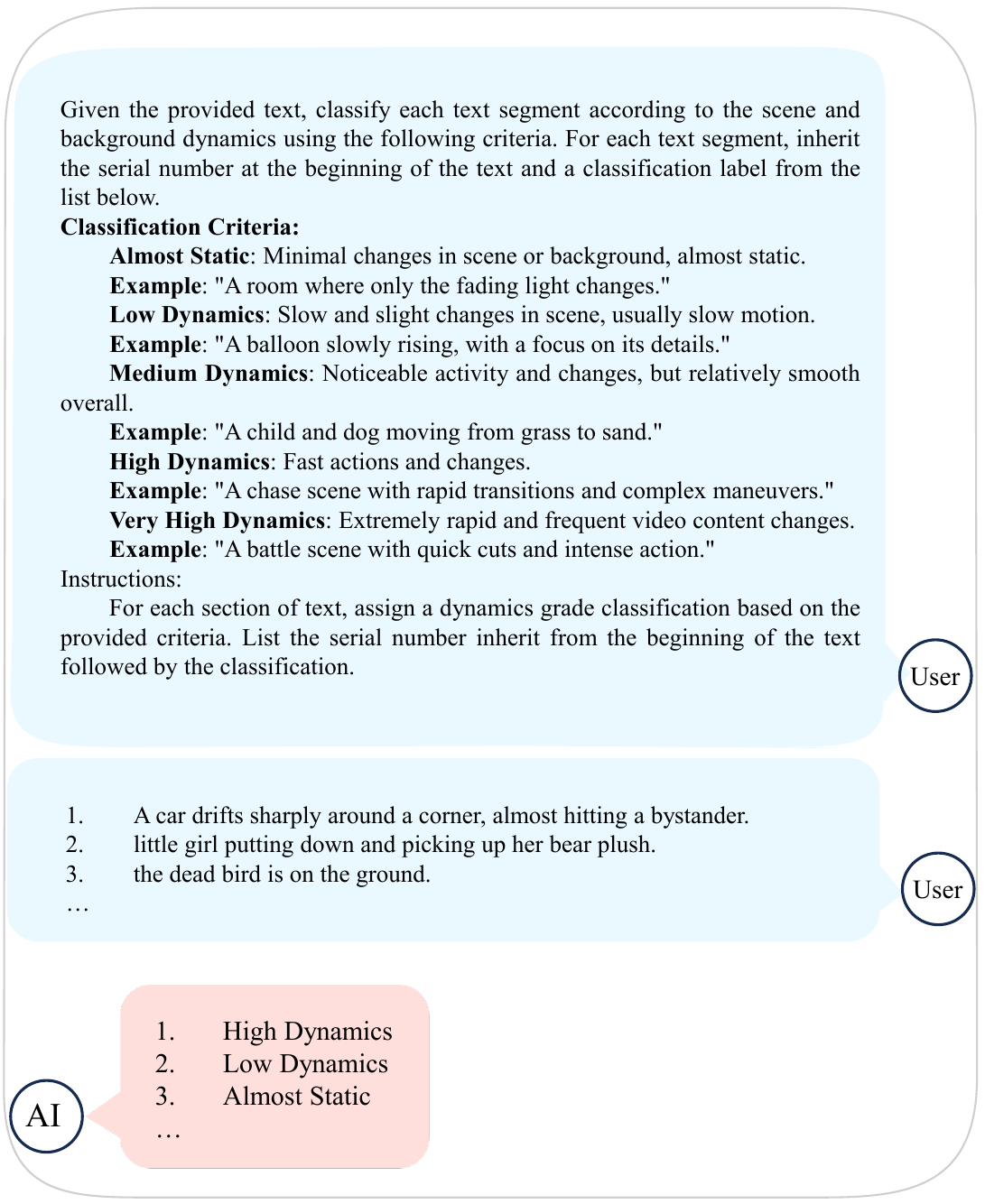}
  \end{center}
 \caption{Illustration of prompt coarse categorization using GPT-4~\cite{openai2023chatgpt}.}
  \label{fig:prompt-categorize-gpt4}
\end{figure}

\begin{figure}
  \begin{center}
    \includegraphics[width=1.0\linewidth]{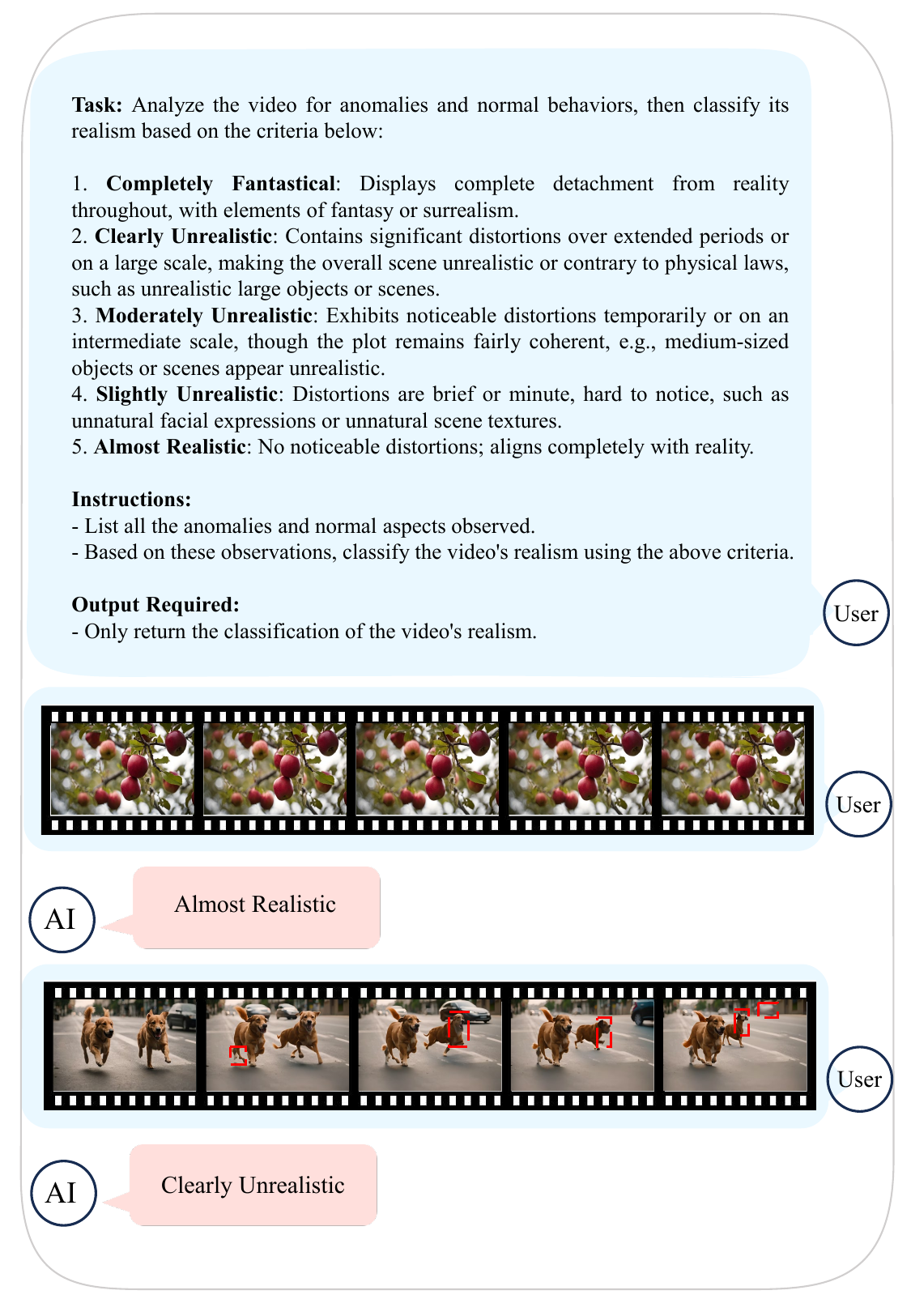}
  \end{center}
 \caption{Illustration of naturalness calculation for generated videos using Gemini-1.5 Pro~\cite{gemini2024}.}
  \label{fig:gemini-naturalness}
\end{figure}

\begin{figure}[htbp]
  \begin{center}
    \includegraphics[width=1.0\linewidth]{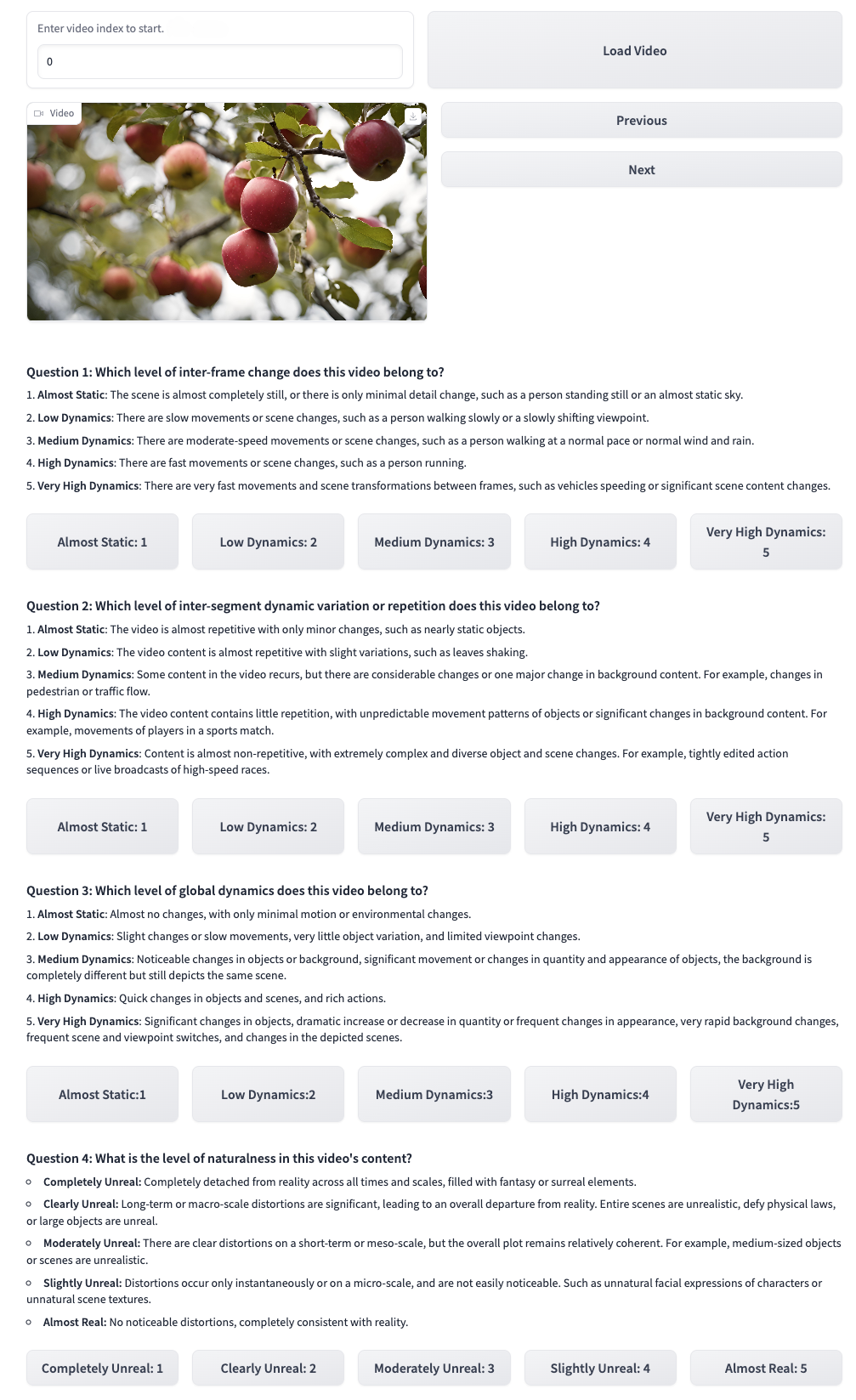}
  \end{center}
 \caption{Toolbox for dynamics and naturalness annotation.}
 \label{fig:ann-interface}
\end{figure}

\section{Visual comparison}

In Section~\ref{sec:dynamics_evaluation_protocol}, we use text prompts with different dynamics grades to generate videos with T2V models. 
Here, we provide visual results of the generated videos.

\begin{figure}[htbp]
  \begin{center}
    \includegraphics[width=1.0\linewidth]{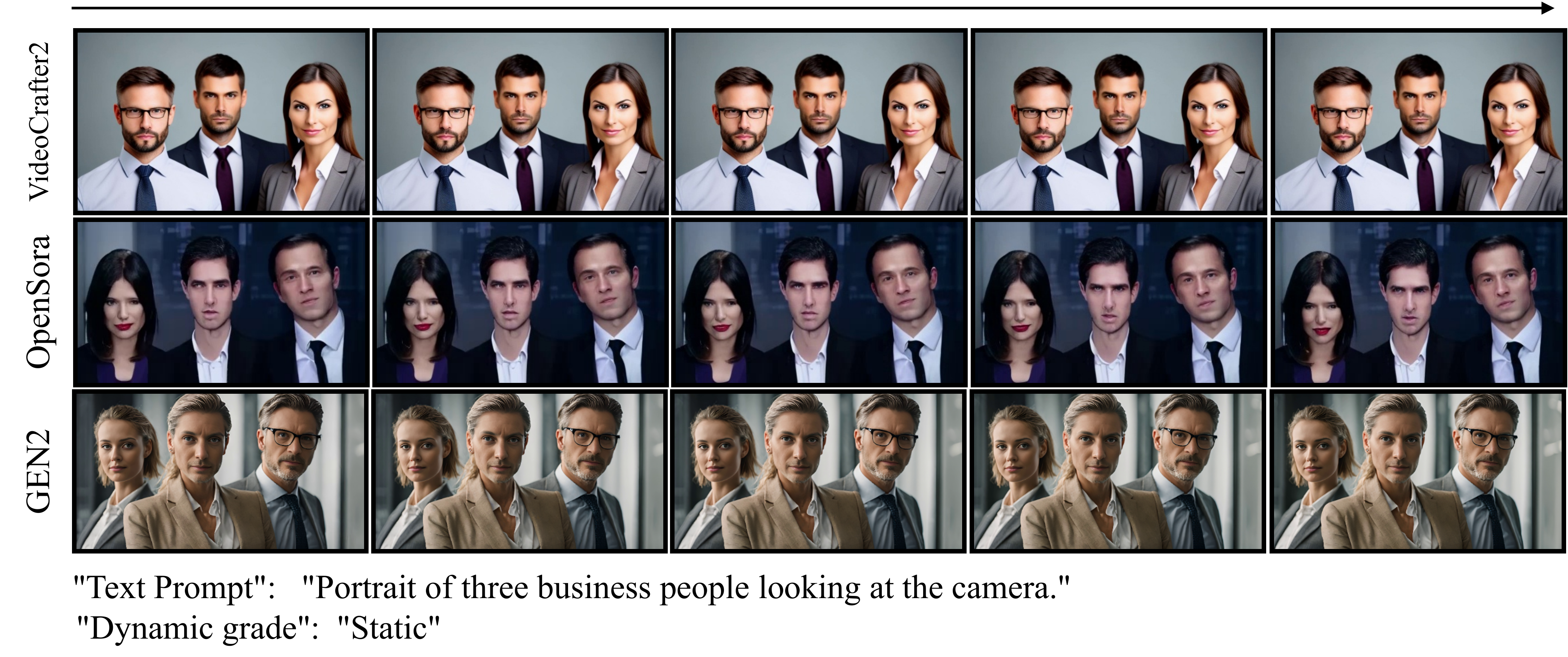}
  \end{center}
\end{figure}

\begin{figure}[htbp]
  \begin{center}
    \includegraphics[width=1.0\linewidth]{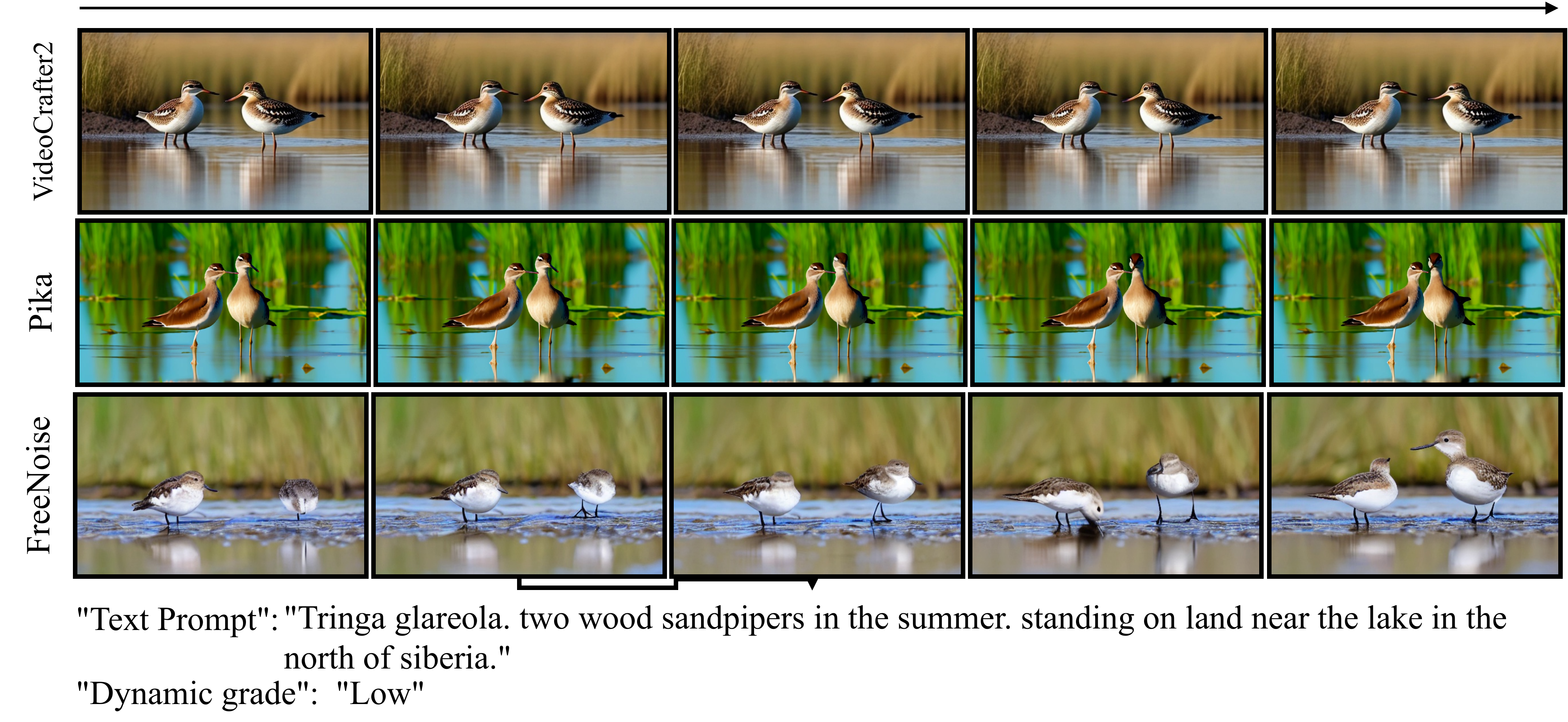}
  \end{center}
\end{figure}

\begin{figure}[htbp]
  \begin{center}
    \includegraphics[width=1.0\linewidth]{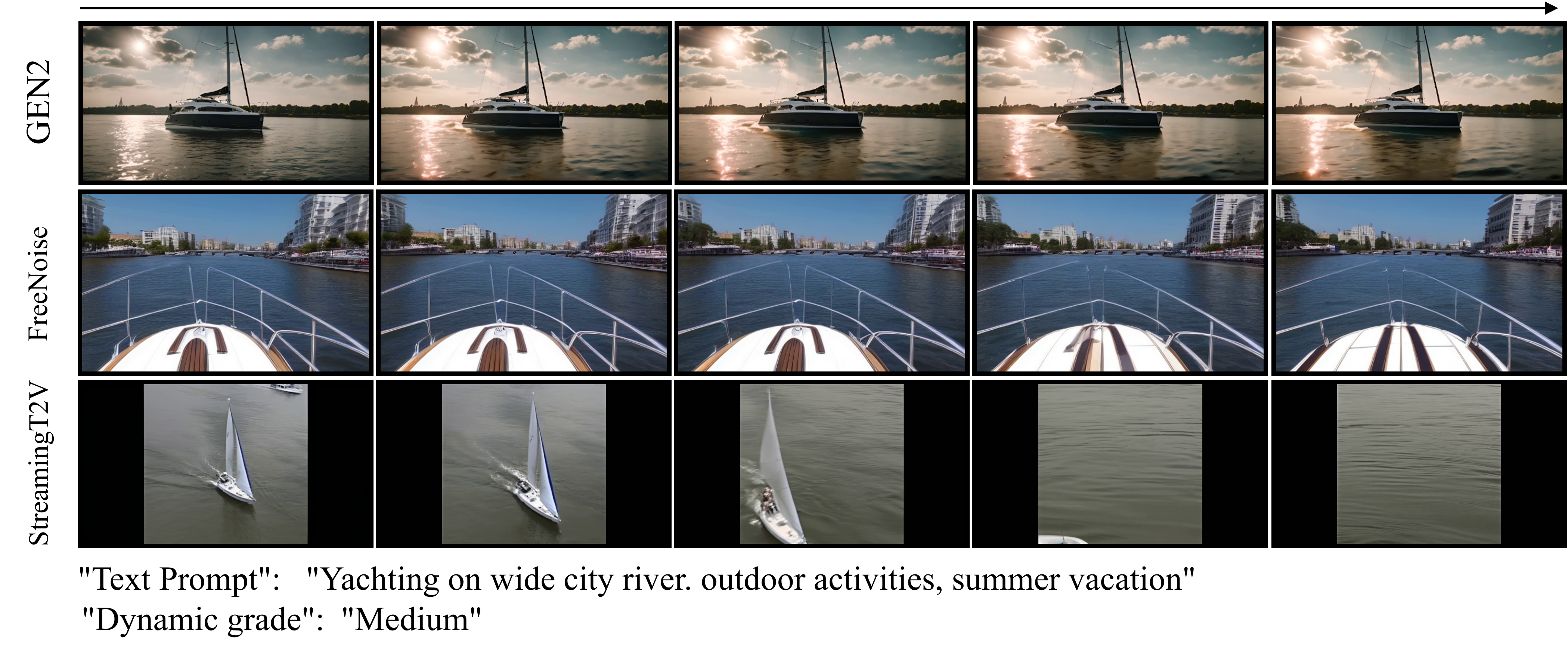}
  \end{center}
\end{figure}

\begin{figure}[htbp]
  \begin{center}
    \includegraphics[width=1.0\linewidth]{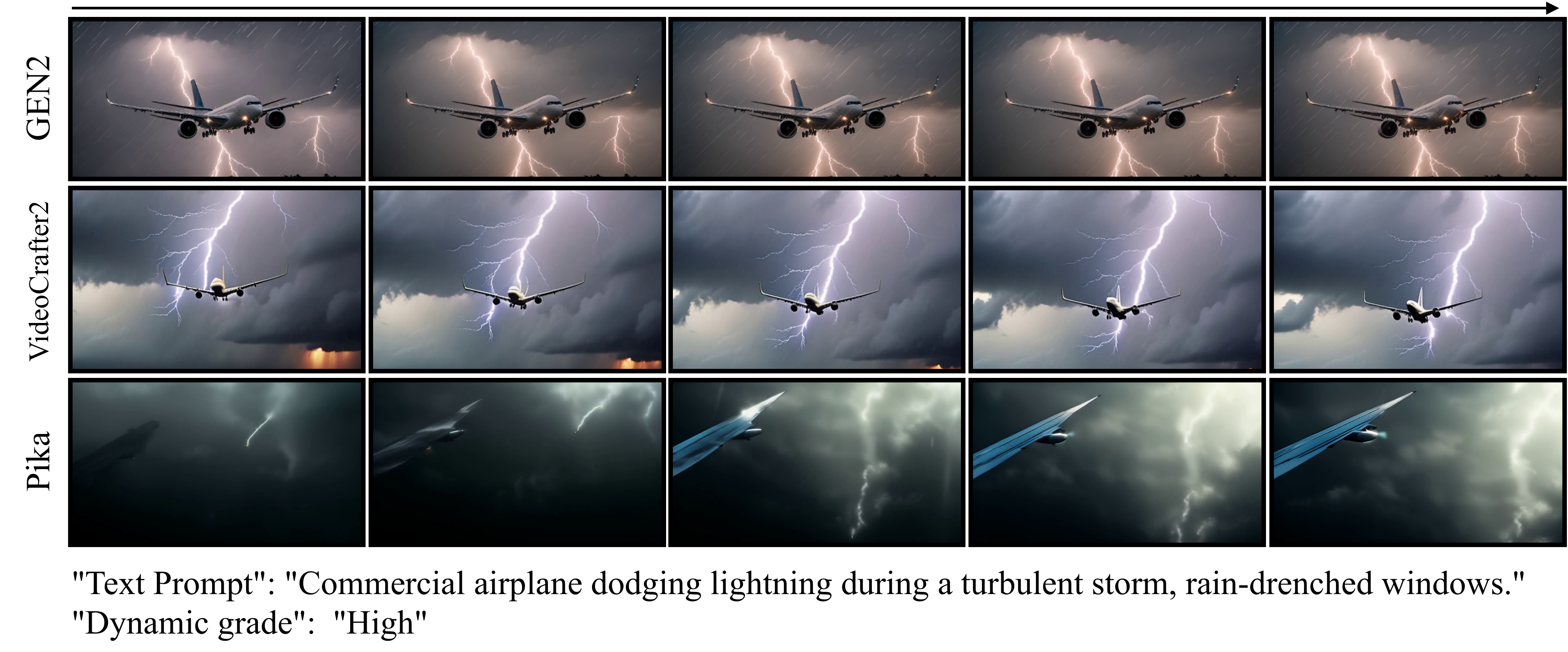}
  \end{center}
\end{figure}

\begin{figure}[htbp]
  \begin{center}
    \includegraphics[width=1.0\linewidth]{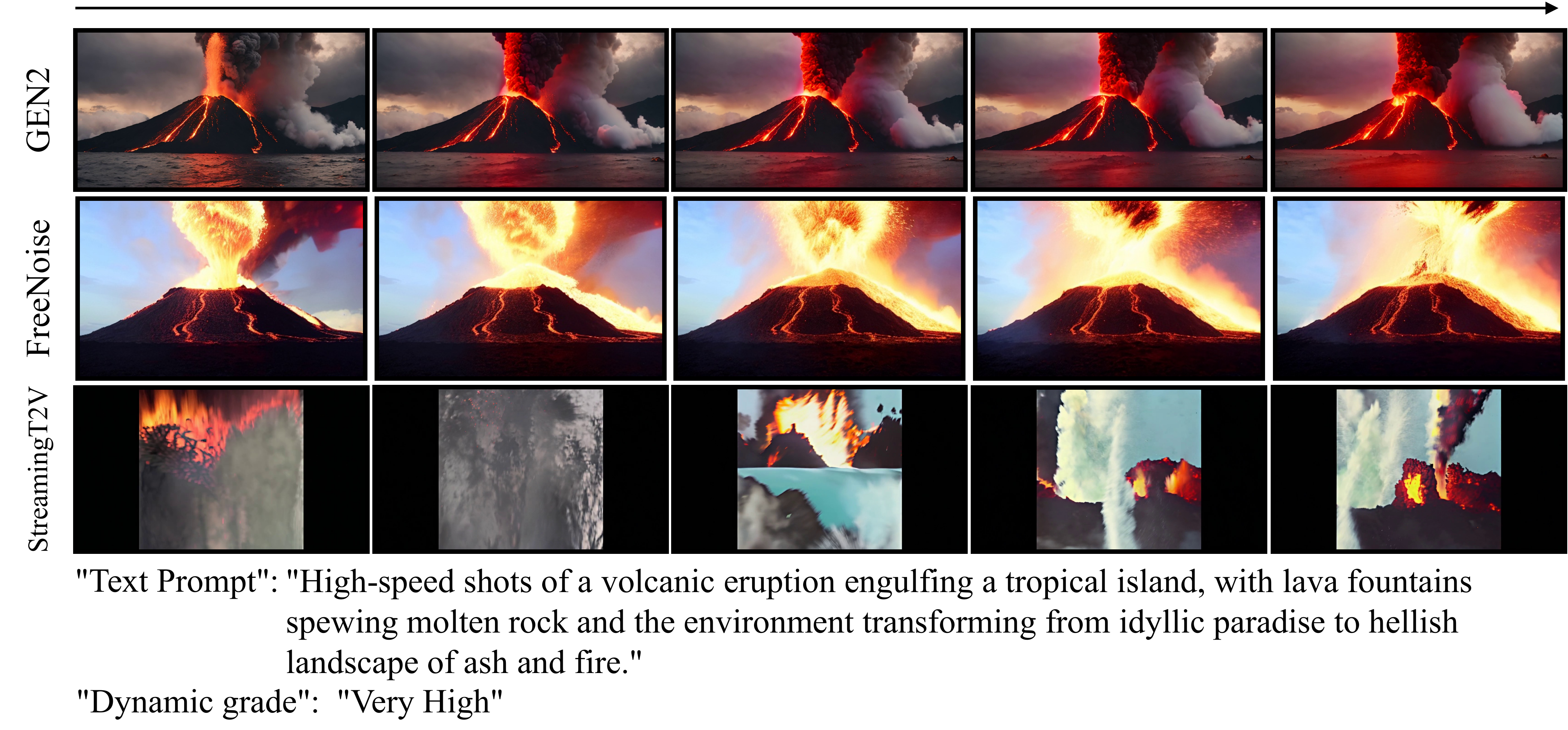}
  \end{center}
\end{figure}

\end{document}